\documentclass{article}

\usepackage[preprint]{neurips_2026}  

\usepackage[utf8]{inputenc} 
\usepackage[T1]{fontenc}    
\usepackage{hyperref}       
\usepackage{url}            
\usepackage{booktabs}       
\usepackage{amsfonts}       
\usepackage{nicefrac}       
\usepackage{microtype}      
\usepackage{xcolor}  
\usepackage{graphicx}
\usepackage{amsmath}
\usepackage{algorithm}
\usepackage{algorithmicx}
\usepackage{algpseudocode}
\usepackage{bm}
\usepackage{multirow}
\usepackage{siunitx}
\usepackage{subfigure}
\title{Not All Tokens Are Worth Caching: Learning Semantic-Aware Eviction for LLM Prefix Caches}

\author{%
  Shaoke Fang\thanks{Equal contribution.} \\
  Peking University \\
  \texttt{fangshaoke@stu.pku.edu.cn} \\
  \And
  Ziang Li\footnotemark[1] \\
  FirestAI \\
  \texttt{ziangli@firestai.com} \\
  \And
  Wenfei Wu \\
  Peking University \\
  \texttt{wenfeiwu@pku.edu.cn} \\
  \And
  Jiatong Ji \\
  Tsinghua University \\
  \texttt{jijt23@mails.tsinghua.edu.cn} \\
  \And
  Qingsong Liu\thanks{Corresponding authors.} \\
  University of Massachusetts Amherst \\
  \texttt{thulqs15@gmail.com} \\
  \And
  Ruizhi Pu\footnotemark[2] \\
  Southeast University \\
  \texttt{rpu2@uwo.ca} \\
}

\begin{document}

\maketitle

\begin{abstract}
    Prefix caching enables the reuse of attention Key-Value (KV) states in Large Language Models (LLM) across requests with shared prompt prefixes and reduces expensive prefill computation. 
    However, its benefits depend critically on the eviction policy as GPU memory is scarce. 
    Existing policies, such as the Least Recently Used (LRU) policies, rely solely on the access time of the cached prefix blocks but treat them uniformly in all other aspects. 
    This uniform view ignores a fundamental property of LLM prompts, namely that, beyond their access patterns (e.g., recency or frequency), not all tokens are equally worth caching given their distinct intrinsic characteristics.
    Empirically, we first observe that different token types within a prompt, including system prompts, user queries, tool outputs, model responses, and chain-of-thought reasoning, exhibit up to 756$\times$ variation in reuse rates, which highlights an interesting yet under-explored justification that the token characteristics matter in prefixing. 
    Motivated by these insights, we present SAECache (Semantic-Adaptive Eviction for prefix caches), a semantic-adaptive prefix cache eviction policy that bridges this gap through three innovations: (1) a multi-queue architecture that routes KV blocks to prefix-block-specific queues with tailored priority metrics, which captures both session reuse in multi-turn interactions and structural reuse in templated single-turn requests; (2) a semantic-aware token weighting mechanism that learns the reuse value of different token types online through eviction feedback; and (3) a fully adaptive online learning schema for all parameters updating, 
    which eliminates manual tuning and enables automatic adaptation to deployment-specific workload characteristics. 
    Through extensive evaluation across heterogeneous workloads, we demonstrate that SAECache achieves a 1.4$\times$ to 2.7$\times$ TTFT improvement over production-style baselines, while showing that fixed-parameter alternatives can severely degrade performance, with a difference of up to 2.7$\times$ worse under workload mismatch, which is a failure mode that our online learning approach avoids entirely.
\end{abstract}







\section{Introduction}
Prefix caching \citep{DBLP:conf/sosp/KwonLZ0ZY0ZS23, DBLP:conf/nips/ZhengYXS0YCKSGB24, DBLP:conf/usenix/GaoHSKJDYYZ24, DBLP:conf/iclr/SrivatsaHAL025, pan2025marconiprefixcachingera} aims to enhance the storage efficiency of Key-Value (KV) states during the inference phase of Large Language Models (LLMs). 
Specifically, the core principle involves leveraging previously computed KV states to mitigate computational redundancy for subsequent requests.
By caching and reusing identical KV states across requests that share input prefixes, prefix caching can significantly reduce computational overhead and thereby improve Time-to-First-Token (TTFT) latency.

However, the benefit of prefix caching is constrained by GPU memory capacity, as it usually occupies scarce GPU memory and can preserve only a small fraction of all reusable KV states.
For example, serving LLaMA-70B on four A100 GPUs leaves room for only about 36K cached prefix tokens. 
Under these circumstances, prefix caching needs to evict certain existing blocks to accommodate new KV states with higher storage values.
Therefore, developing an effective eviction policy for prefix caching has become a first-order design choice: a good policy preserves KV states that are likely to be reused, while a poor one spends memory on blocks with little future value.

\begin{figure}[htbp]
	\centering
	\begin{minipage}{0.325\linewidth}
		\centering
		\includegraphics[width=0.9\linewidth]{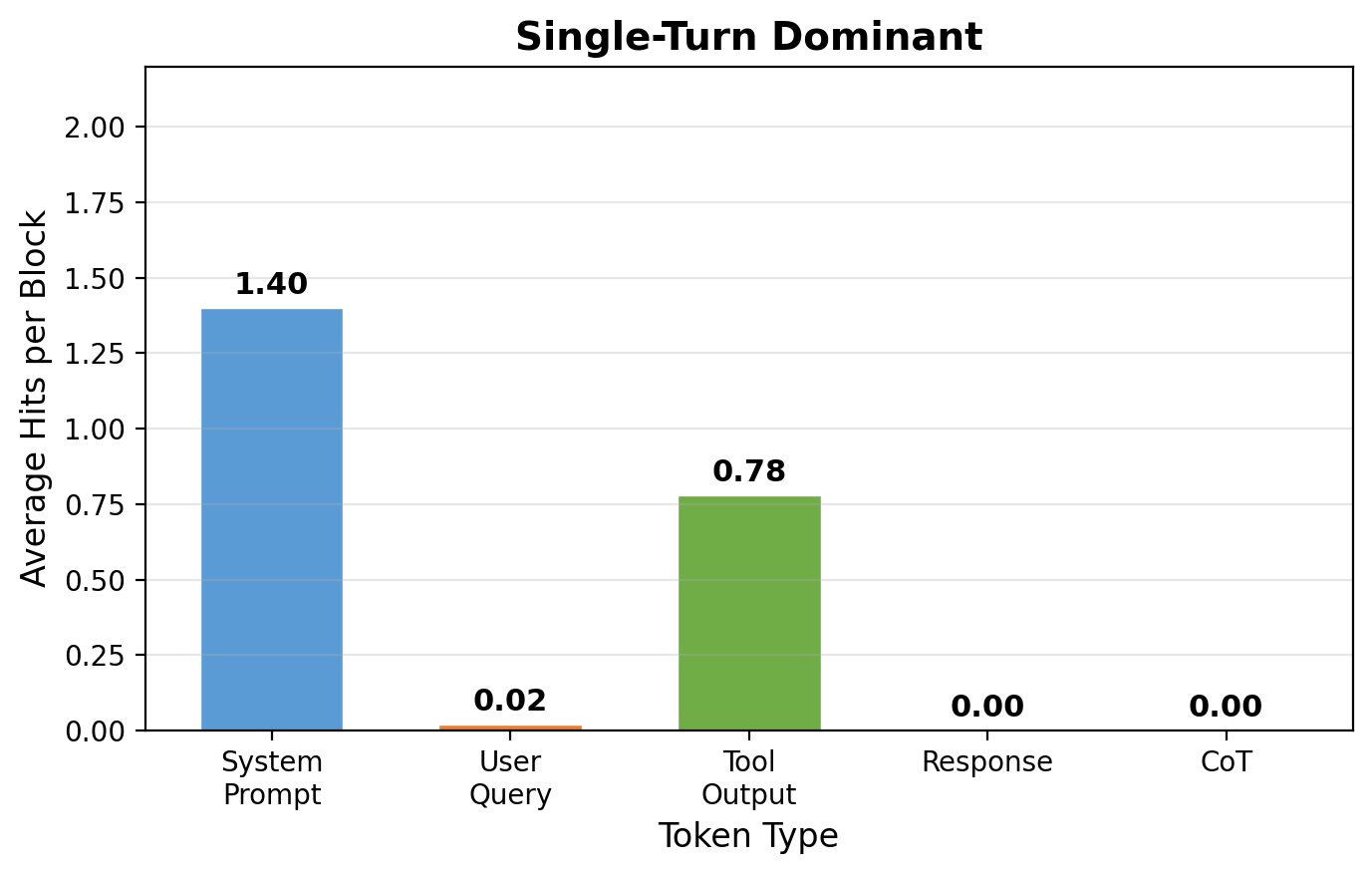}
		\label{single-turn}
	\end{minipage}
	\begin{minipage}{0.325\linewidth}
		\centering
		\includegraphics[width=0.9\linewidth]{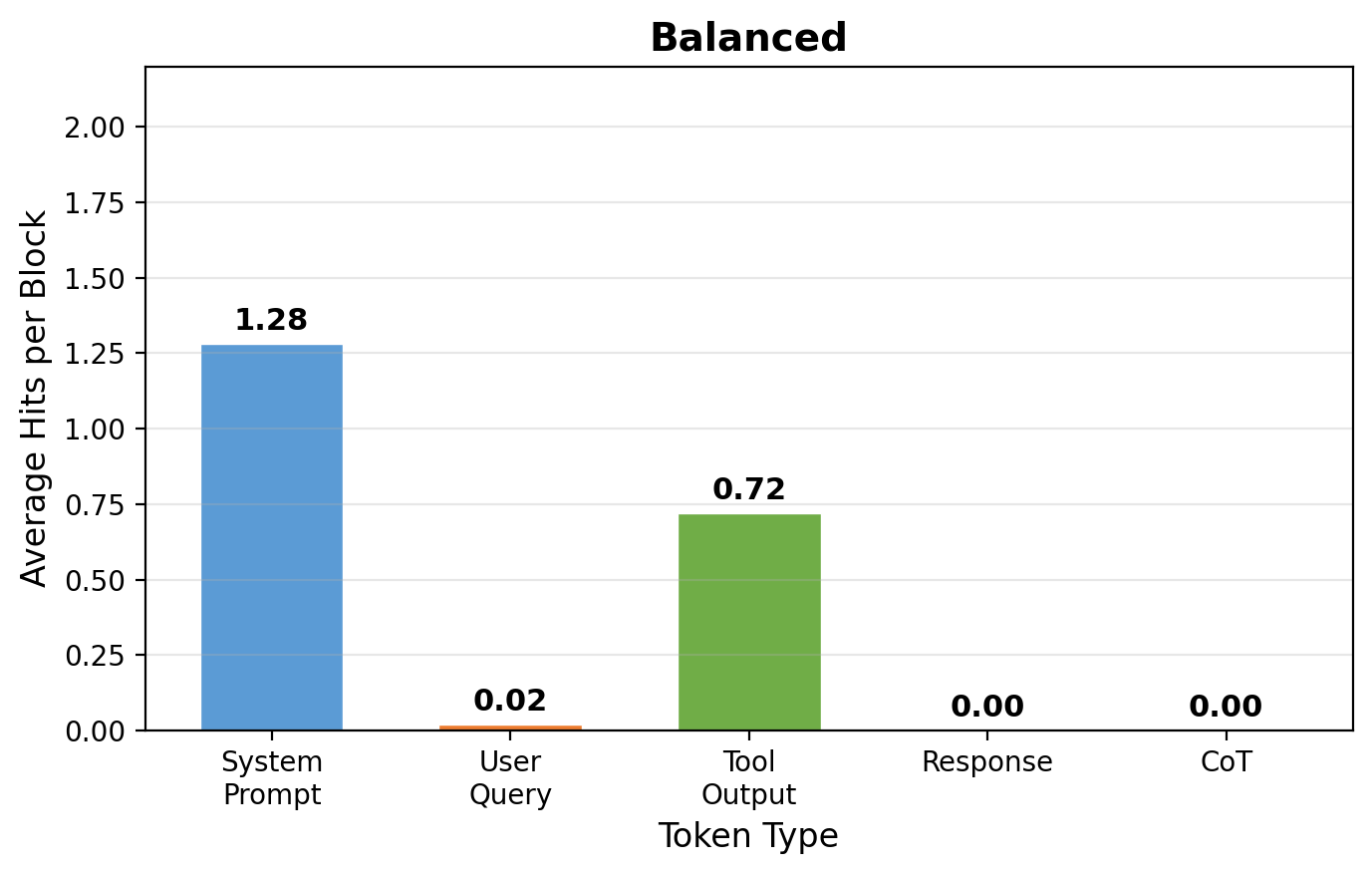}
		\label{balanced}
	\end{minipage}
    \begin{minipage}{0.325\linewidth}
		\centering
		\includegraphics[width=0.9\linewidth]{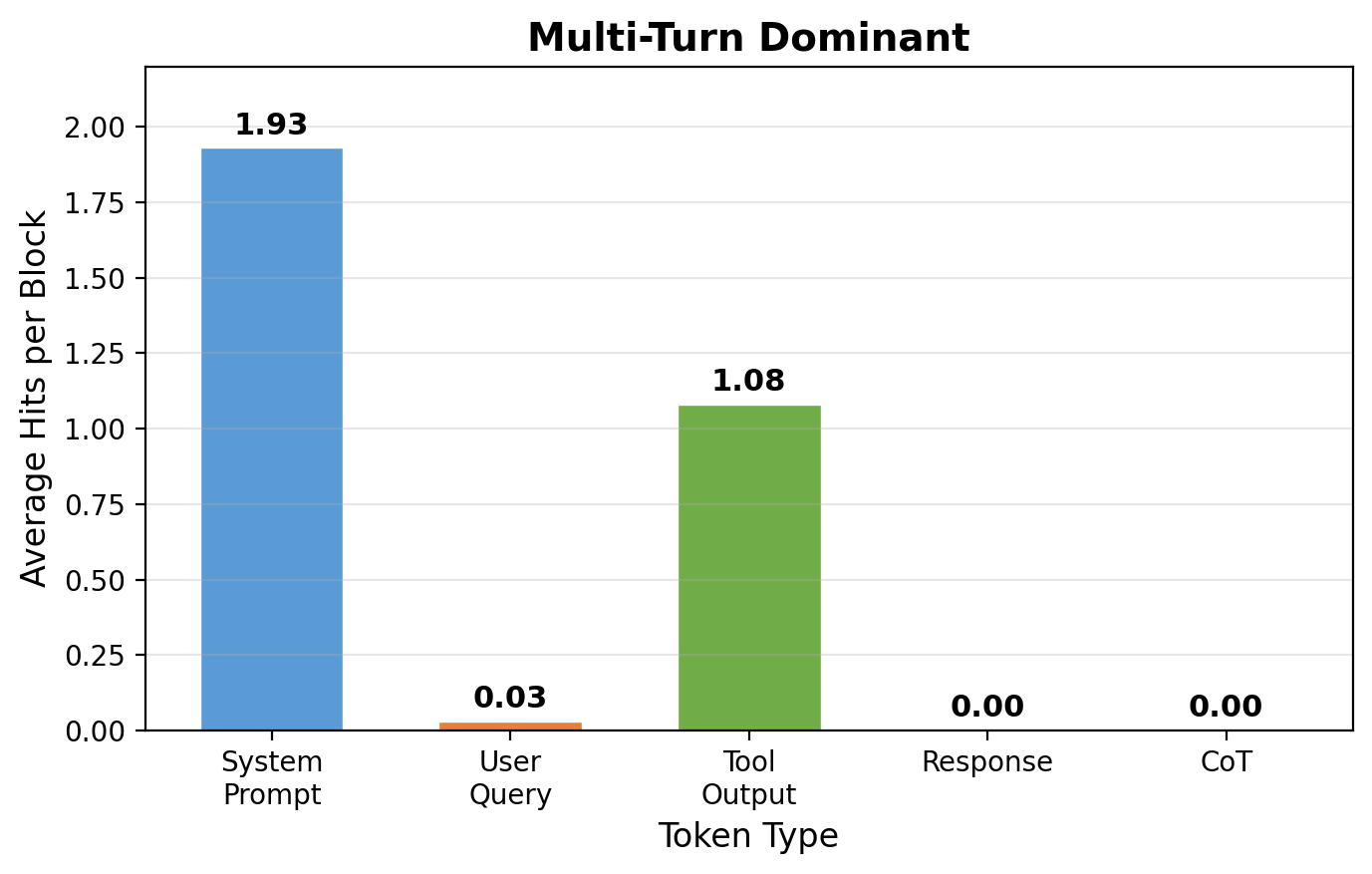}
		\label{multi-tuen}
	\end{minipage}
    \caption{Conversation with different workloads configuration from left to right: 1) Single Turn dominated conversation: 20\% multi-turn + 80\% single-turn tool-use: 100\% single-turn tool-use; 2) Balanced dominated conversation: 50\% multi-turn (chat 30\% + agent 20\%)+ 50\% single-turn; and 3) Multi-Turn dominated conversation: 80\% multi-turn(chat 50\% + agent 30\%) + 20\% single-turn. We can observe a clear distinction between the reuse of different types of tokens.}
    \label{token_resue}
\end{figure}

While existing eviction policies aim to capture only a narrow notion of importance, they remain inadequate for heterogeneous LLM workloads and diverse token semantics. 
Specifically, the Least Recently Used policy (LRU, the default prefix caching policy in vLLM) relies solely on recency for the cached blocks.
However, it ignores both token semantics and workload-specific reuse patterns, which may prematurely evict valuable KV states from multi-turn requests or long-lived sessions with relatively long inter-turn intervals. 
Meanwhile, the Least Frequently Used policy (LFU) suffers from the opposite issue: long sessions may accumulate high access counts, causing stale KV states from finished sessions to linger in cache while evicting useful prefixes from active sessions with fewer access counts. 
Learned Prefix Caching (LPC)~\citep{yang2026learned} is the closest predecessor to our work. 
It enhances LRU by providing predictive guidance to determine which conversations are likely to continue for eviction.
Nevertheless, it remains coarse-grained as all KV blocks associated with a conversation inherit the same continuation signal.
Overall, previous eviction policies are fundamentally limited by their token-agnostic design: they assign uniform importance to all tokens while failing to account for their heterogeneous semantic roles and reuse patterns in practice. We provide a more detailed comparison with related work in Appendix~\ref{app:related_work}.

To better understand the limitations of these previous token-agnostic designs (all existing eviction policies treat all tokens within a prompt essentially uniformly), our characterization through empirical observations reveals why this leaves performance on the table. 
Firstly, as we can observe in Fig. \ref{token_resue} (left), single-turn sessions also exhibit substantial reuse through shared prompt structures, especially in templated tool-use and programming requests. 
Secondly, token semantics can also lead to substantially different levels of reuse in practice: system prompts, user queries, tool outputs, model responses, and chain-of-thought traces can exhibit distinct reuse patterns (As we show in Fig.\ref{token_resue} and Section~\ref{sec:prefixreuse}, the reuse rate varies by as much as 44× across token types, from 92.3\% for system prompts to only 2.2\% for chain-of-thought tokens).  
Therefore, treating a system-prompt block and a reasoning-trace block as equivalent whenever their recency statistics match can result in suboptimal memory allocation.
Lastly, the reuse in multi-turn sessions is almost entirely session-local, but the timing of that reuse differs sharply across different workloads (e.g., chat and agentic).
What's more, we observe that inter-turn intervals in each session can be well captured by log-normal distributions, with their parameters varying across different sessions.

Motivated by these observations, we propose SAECache, a semantic-aware eviction policy for heterogeneous LLM serving that integrates token-semantic structure with online workload adaptation. 
The core idea is to move beyond fixed, token-agnostic heuristic eviction policies and let the cache learn to evict cached prefix blocks in a fine-grained manner.
We aggregate the token characteristics and real-time serving feedback to construct a weighted eviction score to determine which content is least valuable to evict.
Our contributions are fourfold and can be summarized as follows.

\paragraph{Our Contributions}
Firstly, we present a systematic empirical study of prefix reuse across diverse session structures and different token-type granularity levels.
We observe that single-turn sessions create mainly templated structural reuse, while the token-type reuse varies by more than 10 times compared to each other in multi-turn sessions.
Meanwhile, beyond confirming the session locality as that of \cite{yang2026learned}, we show that inter-turn intervals follow session-specific log-normal distributions (Section~\ref{sec:prefixreuse}).

Secondly, motivated by our empirical observations, we propose SAECache, a semantic-aware multi-queue eviction policy that leverages a rule-based partition method to assign cached KV states into multiple prefix-block-specific queues to process the requests.
Specifically, we introduce four distinct queues that represent different types of prefix blocks in the cached KV states: (1) an evict-first queue that stores KV states associated with low-value, untemplated, and decode-phase blocks; (2) a structural queue that stores KV states associated with templated structured and templated blocks; (3) an agentic queue that stores KV states associated with agentic and tool-use blocks; and (4) a chat queue that stores KV states associated with multi-turn tokens (Section~\ref{sec:saecache}).
Moreover, to facilitate the queuing process and enhance the queuing quality, we introduce a learning-guided predictor that identifies incoming LLM requests as either single-turn or multi-turn sessions considering their different reuse potentials.

Thirdly, to enable comparable eviction priorities across queues, we introduce an online learning-guided dynamic token-type-aware weighting mechanism to build a priority metric for eviction in the four prefix-block-specific queues.
Notably, we make all major token-type-aware parameters adaptive. 
In particular, it estimates the log-normal timing parameters for chat and agentic queues via online maximum likelihood estimation (MLE), learns the structural queue’s position-decay parameter from hit-rate correlations using exponential moving average (EMA) smoothing, and updates token-type and queue weights according to eviction feedback like hit efficiency (Section~\ref{sec:saecache}).

Lastly, we conduct extensive experiments on real-world datasets, demonstrating that SAECache improves TTFT by 1.4$\times$ to 2.7$\times$ on heterogeneous workloads. 
Ablation studies further show that each learning component contributes meaningfully to the overall gain: token-type-aware weighting yields an additional 23\% improvement, while adaptive queue weighting contributes 39\%. 
Importantly, we also show that fixed-parameter alternatives can degrade by up to 2.7$\times$ under workload mismatch, which highlights the necessity of online adaptation in practical LLM serving systems (Section~\ref{sec:exp}).

\section{Prefix Reuse Characterization and Design Insights}
\label{sec:prefixreuse}
This section highlights key observations in prefix reuse that drive our design.
Prior work, LPC \citep{yang2026learned} showed that multi-turn conversations create valuable prefix reuse and that recency alone is insufficient. 
We revisit that setting and add three missing dimensions: structural reuse in single-turn prompts, semantic reuse heterogeneity within a prompt, and distributional inter-turn timing. 
These measurements and observations motivate an eviction policy that is semantic-aware, prefix-block-specific, and adaptive.

\subsection{Single-Turn Reuse Comes From Prompt Structure}

\begin{figure}
    \centering
    \includegraphics[width=0.95\linewidth]{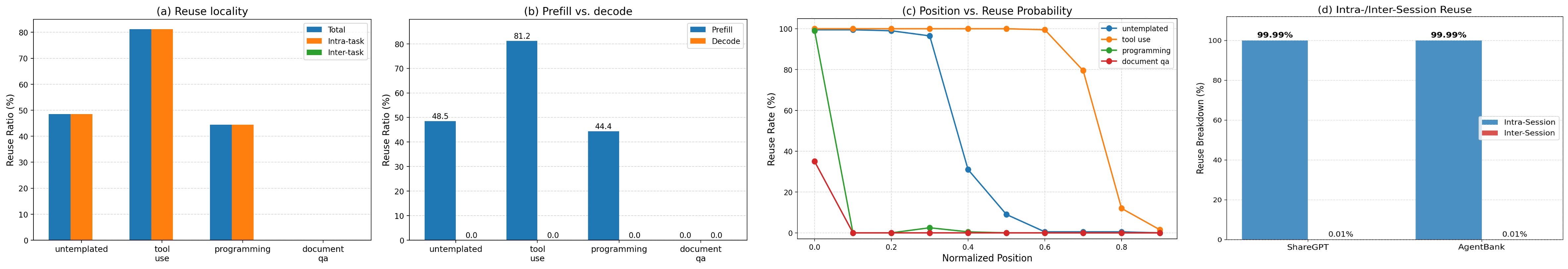}
    \caption{a) the reuse locality breakdown with overwhelming session-local dominance and negligible cross-task reuse; b) the hit ratio comparison confirming the superiority of tail-first eviction; c) impact of token position for evictions on KV cache reuse. 
    Illustration on the mechanism that evicts from the tail preserves structural prefix sharing while evicting from the head breaks it; and d) Intra-/inter-session reuse breakdown in multi-turn sessions. Bar chart showing more than 99\% intra-session reuse for both ShareGPT and AgentBank, with negligible inter-session reuse.}
    \label{prefix_emp}
\end{figure}

Single-turn requests typically come from a stateless API call session. 
Although such sessions may initially appear unfavorable for prefix caching (due to a lack of conversational history), they still exhibit potentially substantial reuse due to shared prompt structure (Fig. \ref{prefix_emp} a and b). 
For example, tool-use prompts often share a system prompt and tool descriptions; programming workloads reuse fixed instructions and session framing; and document QA may share a stable request wrapper even when the document changes. 
This reuse is session-local: cross-session reuse is below 0.01\% because different applications use different templates.

Specifically, we identify two additional patterns that shape single-turn eviction. 
First, decode blocks have reuse rates below 3.17\% since reuse would require an exact match of generated tokens.
Therefore, they should be evicted before scarce capacity is taken from reusable prefill prefixes. 
Second, prefill reuse decays with the prompt position from the requests \ref{prefix_emp} c. 
Since prefix matching is strict, a later block can be reused only when every preceding token also matches. 
Earlier blocks that contain system instructions or shared scaffolding are therefore more valuable than later blocks that contain request-specific content.

\paragraph{Design insight.}
The reuse of the Prefix cache blocks from single-turn sessions mostly consists of templates, and these templated single-turn blocks should be scored by position, with earlier blocks receiving higher priority. 
The decay strength should be learned because the template length and amount of shared structure vary across applications.

\subsection{Multi-Turn Is Session-Local and Distributional}

In multi-turn sessions, since each new request usually contains the preceding conversation history, the reuse of the prefix blocks is primarily driven by session continuity. 
Specifically, this reuse is almost entirely intra-session: 99.28\% of reused blocks in ShareGPT-style chat traces and 99.96\% in AgentBank-style agentic traces come from the same session. 
Cross-session reuse is negligible because exact token-level prefix matches rarely occur across independent conversations (Fig. \ref{prefix_emp} d).

\begin{figure}
    \centering
    \includegraphics[width=0.9\linewidth]{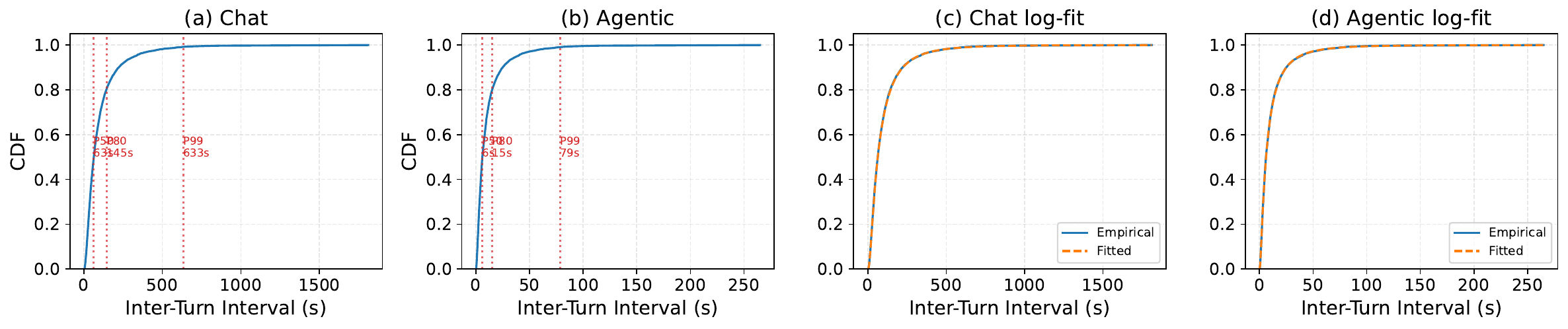}
    \caption{Inter-turn interval Cumulative Distribution Functions (CDF) for chat and agentic workloads. Panels (a) and (b) show empirical distributions with distinct timescales, while panels (c) and (d) show log-normal fits that closely track the empirical distributions.}
    \label{fig:interturn_cdf}
\end{figure}

Meanwhile, we can observe that the arrival times of the requests (request interval) across the sessions vary significantly.
Specifically, each request interval of each session follows a log-normal distribution with distinct hyperparameters.
We find that these intervals vary substantially by workload: human-facing chat has long pauses with a heavy tail spanning two orders of magnitude (P50\footnote{P50 denotes that at least 50\% of the data points are less than or equal to the P50 value.}$\approx$110\,s, P80$\approx$372\,s, P99$\approx$2{,}207\,s in Qwen-Bailian), while agentic traces are an order of magnitude burstier (P50$\approx$8.5\,s, P80$\approx$25\,s in CC-Bench). Both, however, are well-fit by log-normal distributions ($R^2\geq 0.987$, K-S $\leq 0.060$, see Appendix~\ref{app:lognormal-params}).
Despite the different time scales across the sessions, both can be well approximated by log-normal distributions (Figure~\ref{fig:interturn_cdf}). 
This is an important departure from LPC's \cite{yang2026learned} manually tuned exponential decay, whose scale is chosen from an average turn interval (e.g., 0.01s). 
Beyond them, a distributional model in our work captures the entire reuse timing curve and can be estimated online for each deployment.

\textbf{Design insight.} 
Cached prefix block reuse in multi-turn sessions is session-local due to the conversation history.
Blocks that have remained unused far beyond the learned distribution are less likely to be reused soon and should lose priority
The request intervals across the sessions vary significantly, but each follows a unique log-normal distribution, and the parameters should be learned online.
Moreover, single-turn sessions have different reuse potential compared to multi-turn sessions, as the reuse of former blocks mostly falls in the structural templates, while the content tokens in the latter would also potentially be reused.

\subsection{Token Types Have Unequal Reuse Value}
\label{sec:2.3tokentypes}
\begin{figure}[H]
    \centering  
    \includegraphics[width=\textwidth]{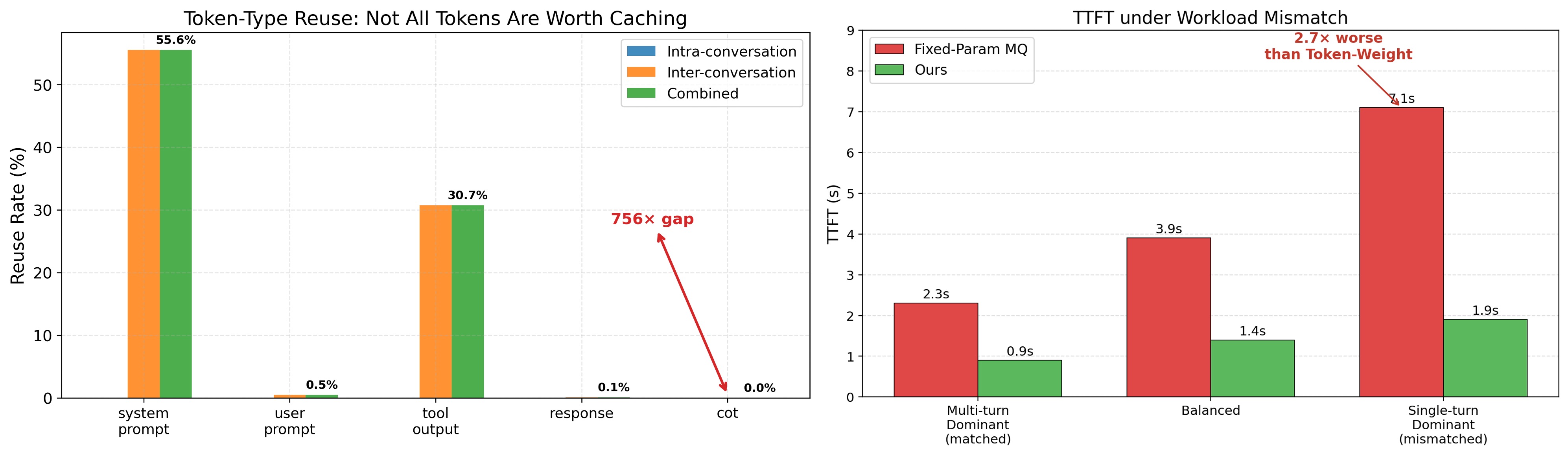}
    \caption{Left)Token-type reuse rates. Grouped bar chart showing intra-conversation and inter-conversation reuse rates for each token type, revealing the 756$\times$ variation between system prompts at 92.3\% and chain-of-thought tokens at 2.2\%; and Right)TTFT under workload mismatch. Fixed-parameter performance degrades as workload composition deviates from the trace used for fitting, increasing from 4.014 seconds on the matched workload to 7.127 seconds on the mismatched workload, while LearnedCache maintains stable performance, ranging from 0.934 seconds to 1.886 seconds, across all compositions. }
    \label{type_reuse_ttft}
\end{figure}

Beyond the session structures, LLM requests typically contain semantically distinct tokens, such as system prompts, user queries, tool outputs, model responses, and chain-of-thought traces. T
These tokens serve different roles in the prompts and have distinct future values, which motivates us to measure reuse by token type.
As we can observe from Table~\ref{tab:token_type_reuse} and Fig.\ref{type_reuse_ttft} Left, system prompts have the highest reuse because they encode stable application instructions. 
User prompts have moderate reuse from structured query templates offset by variable per-request content. 
Response tokens exhibit a notable asymmetry phenomenon: high intra-conversation reuse (36.3\%) as multi-turn prompts include previous historical responses, yet low inter-conversation reuse (1.5\%) since different sessions contain different semantic content in requests and responses.
Chain-of-thought tokens are the least reusable because reasoning traces are highly query-specific and can vary significantly even for similar inputs due to stochastic decoding in LLM.

\begin{table}[t]
  \caption{Reuse rates by token type. System prompts exhibit the highest reuse (92.3\%), while chain-of-thought tokens are rarely reused (2.2\%), revealing a $42\times$ variation in combined reuse rate.}
  \label{tab:token_type_reuse}
  \centering
  \begin{tabular}{lccc}
    \toprule
    Token Type & Intra-conv.\ (\%) & Inter-conv.\ (\%) & Combined (\%) \\
    \midrule
    System prompt    & 96.0  & 85.7  & 92.3 \\
    User prompt      & 43.3  & 13.6  & 30.8 \\
    Response         & 36.3  &  1.5  & 27.8 \\
    Tool output      & 31.2  & 12.2  & 23.0 \\
    Chain-of-thought &  4.2  &  0.0  &  2.2 \\
    \bottomrule
  \end{tabular}
\end{table}

\paragraph{Design insight.}
Eviction should be semantic-aware by tokens. 
A system-prompt block and a chain-of-thought block should not receive the same priority, even when their recency or frequency statistics match. 
Token-type weights should be learned from feedback since workload composition can shift from time to time.

\subsection{Additional observation: Fixed Parameters Are Fragile}
Based on previous observations, we also explain that fixed policies are brittle in dynamic LLM systems.
A policy fitted to chat-heavy traffic can learn timing and queue weights that are inappropriate for single-turn-dominant traffic. 
In our empirical study, log-normal parameters fitted from a chat-dominated trace ($\mu=4.15$, $\sigma=0.97$) work well for chat-like workloads but cause 2.7$\times$ TTFT degradation when applied to single-turn-dominated workloads (in Fig.\ref{type_reuse_ttft} Right). 
Fixed queue weights that prioritize session queues at a ratio of 24:1 over structural queues catastrophically misallocate cache capacity when structural reuse dominates.

\paragraph{Design insight.}
Parameters and weights should be updated online from feedback (e.g., hit efficiency) as the relative reuse value of different token types is workload-dependent and can change over time. 
A static weighting scheme may over-prioritize token types that were useful in past workloads but provide limited future cache hits under the current workload.
This includes timing distributions, position decay, token-type weights, queue weights, and the smoothing/temperature parameters that govern adaptation speed.

\section{SAECache: Semantic-Aware Prefix Eviction}
\label{sec:saecache}
Building on the above observations and analysis, we propose SAECache, a semantic-aware eviction policy that translates these design insights into a practical online learning algorithm to effectively improve prefix caching efficiency.
This policy consists of three components: queue routing, local priority scoring, and adaptive parameter learning.

\subsection{Queue Routing, Local priority scoring and adaptive parameter learning}

\textbf{Queue routing structural}
SAECache routes each KV block to the queue based on its dominant reuse patterns from our observations, regardless of whether they are from single- or multi-turn sessions.

\begin{itemize}
    \item \textbf{Evict-First Queue:} holds decode blocks and low-reuse untemplated prefill blocks. This queue is drained before any other queue.
   
    \item \textbf{Structural Queue:} contains templated single-turn prefill blocks. This queue captures reuse from the shared prompt structure.
    
   \item \textbf{(Multi-turn) chat queue:} maintained separately for chat multi-turn blocks. These queues model session-local temporal reuse.

   \item \textbf{(Multi-turn) agentic queue:} maintained separately for agentic multi-turn blocks. These queues also model session-local temporal reuse, but differ in token content and request interval across sessions compared to the chat queue.
\end{itemize}
The overview of our multi-queue framework can be found in Appendix \ref{adaptive} Fig.\ref{fig:framework_overview}

\textbf{Local priority scoring.} 
Since prefix blocks reuse in single-turn sessions mainly arises from prompt templates, these blocks are directed to either the evict-first queue or the structural queue. 
To improve reuse efficiency, we primarily focus on prefix block reuse in multi-turn sessions.
Therefore, we propose a local priority scoring mechanism to score each block for eviction.

For a block $b$ in a multi-turn queue $q$, let $\Delta t_b$ be the elapsed time since its last access. 
Its local survival priority is the survival probability under the queue's learned log-normal model \footnote{The learning of such a log-normal model can be found in Appendix \ref{app:lognormal-params}}:
\begin{equation}
p_q(b) = 1 - F_{\mathrm{LN}}(\Delta t_b;\,\mu_q,\sigma_q),
\end{equation}
where $F$ denotes the CDF of the fitted log-normal distribution.
A larger value means the block is more likely to be reused soon and should be retained. 
We use separate $\mu_q^c,\sigma_q^c$ for chat and $\mu_q^a,\sigma_q^a$ for agentic queues as their request intervals differ substantially (e.g., the human chat would exhibit a longer request interval than the agent applications).

For a block $b$ in the structural queue, let $o_b$ be its block offset and $o_{\max}$ the maximum offset in the request. 
The local survival priority is written as:
\begin{equation}
    \label{position_rule}
    p_q(b) = 1 - \left(\frac{o_b}{o_{\max}}\right)^{\gamma},
\end{equation}
where $\gamma$ is the decay power that is learned online, and more importantly, earlier blocks can receive a higher retention priority.

Instead of directly comparing the local survival priorities from different queues, SAGECache incorporates two additional fine-grained token-semantic-aware information for eviction.
First, $w_{\tau(b)}$ denotes the token-type weight for the block $b$, where we set different initial weights for the different types of tokens.
Meanwhile,  since the block size is much smaller than the prompt size (e.g., 16 tokens per block), there is a high probability that tokens in the same block share the same token types (e.g., a system prompt contains hundreds of tokens and is larger than 16).
Therefore, we select the type of median token in the block to represent the block type $\tau(b)$, and we use the miss-after-eviction feedback to increase the weight if hit and vice versa.
Second, we introduce the queue-level hit-efficiency weight $\alpha_q$ to measure the recent hit efficiency of the queue, which can accurately reflect the dynamics of our eviction effectiveness. 
In summary, the local priority score can be calculated by:
\begin{equation}
P(b) = \alpha_q \cdot w_{\tau(b)} \cdot p_q(b) \; / \; \Delta t_b,
\end{equation}

\textbf{Eviction procedure.}
SAECache \emph{first} evicts blocks from the evict-first queue. 
If this queue is empty, it evicts the block with the smallest score sorted among those in the multi-turn and structural queues.
Algorithm \textbf{SAGECache~Eviction} (in Appendix \ref{alg:global_eviction}) shows the detailed eviction procedure.
This procedure lets multi-turn session timing, single-turn prompt position, token semantics, and queue-level utility to jointly determine eviction. Figure~\ref{fig:framework_overview} illustrates the eviction workflow of SAECache.

\textbf{Adaptive Parameter Learners.}
All parameters are learned online through a unified feedback loop: token-type weights, log-normal parameters $(\mu, \sigma)$, position decay power $\gamma$, queue weights $\alpha$, and meta-parameters such as temperature $T$ and exponential moving average (EMA) smoothing factor $\beta$. 
The update rule for each parameter is lightweight and uses only queue-level counters collected during normal cache operation.
The online overhead is constant per eviction, apart from candidate selection, which can be implemented with per-queue heaps. 
The details of parameter initialization, updating rules, and their corresponding running costs can be found in the Appendix \ref{adaptive}.

\subsection{Multi-turn Session Predictor}
\label{sec:predictor}
However, a subtle routing challenge arises at the first request of a conversation.
If a completed request already carries conversation history, SAECache treats it as part of an ongoing multi-turn session and routes its reusable prefix blocks to the corresponding multi-turn queues (depending on whether it is a chat or agentic type).
If the request has no history, however, it is ambiguous: it may be a stateless single-turn request, or it may be the first request of a multi-turn session. SAECache resolves this ambiguity with a lightweight multi-turn session predictor (Section~\ref{sec:predictor}). 
Specifically, history-free requests predicted to continue are routed to a multi-turn queue, templated requests are routed to the structural queue, and untemplated low-reuse blocks from both single and multi-turn sessions would directly fall back to the evict-first queue.
Since in the standard vLLM implementation \cite{DBLP:conf/sosp/KwonLZ0ZY0ZS23}, the first request is set to be the single-turn session by default.
Considering that multi-turn sessions have higher reuse potentials than single-turn sessions, it is crucial to identify whether a request belongs to a single or multi-turn session before routing it to its corresponding queue.

Therefore, we propose a learning-guided session predictor to identify the belonging of the first request of a conversation.
Our lightweight predictor operates only on the serving model's own hidden representation. 
Since the hidden state of the last input token integrates the information from all the previous tokens through the attention mechanism, it has been widely adopted in downstream LLM tasks \cite{chen-etal-2023-token} (e.g., classification).
During prefill, we employ the final-layer hidden state $\mathbf{h}\in\mathbb{R}^d$ of the last input token and pass it to a three-layer MLP with $W$ denoting the layer weights and $b$ denoting the bias:
\begin{equation}
\hat{y} = W_3 \sigma(W_2 \sigma(W_1 \mathbf{h} + b_1) + b_2) + b_3,
\end{equation}
where $\sigma$ denotes ReLU activation and $\hat{y}\in \{0,1\}$ is a binary prediction indicating whether the request is likely to become multi-turn. 
The MLP uses hidden dimensions of 256 and 64, totaling roughly one million parameters. 
Crucially, this MLP reuses features that the serving model has already computed during prefill, and does not require a separate encoding pass or an external model.

\section{Experiments}
\label{sec:exp}
This section evaluates the efficiency of our proposed policy in terms of cache hit ratio and latency.


\textbf{Baselines and workloads.}
We compare SAECache against two main baselines: LRU Baseline, a production-style default policy used in systems such as vLLM \cite{chen-etal-2023-token}, and LPC \cite{yang2026learned}, the closest predecessor to our work.
We evaluate all policies on three representative real datasets with different levels of multi-turn traffic: ShareGPT, where $74\%$ of requests belong to multi-turn sessions; LMSys, where $33\%$ of requests are multi-turn; and Chatbot-Arena, where only $12\%$ of requests are multi-turn.
These datasets allow us to evaluate eviction policies across workloads ranging from multi-turn-session-dominant to mostly single-turn traffic.
For the ablation study in Appendix~\ref{ablation_study}, we additionally consider different variants of SAECache for making a comprehensive comparison.

Also, since prefix cache hits directly reduce the number of tokens that require prefill computation, TTFT (time-to-first-token) is the primary latency metric affected by cache eviction.
Specifically, cache hits reduce the effective prefill length as
$
\text{prefill\_tokens}
=
\text{prompt\_length} \times (1 - \text{hit\_ratio}).
$
We therefore report the mean TTFT and prefix cache hit ratio for all compared policies.

\textbf{Prototype implementation on vLLM.} 
To verify deployability in a real LLM serving system, we implement SAECache and baselines as drop-in replacements for vLLM's prefix cache evictor module. 
The SAECache implementation adds a new Cache Evictor class that replaces the default LRU Evictor in vLLM. 
This integration requires no changes to vLLM's scheduler, model executor, or serving infrastructure
We verify the integration on vLLM v0.8.5 (V0 engine), using Qwen2.5-1.5B-Instruct on an NVIDIA A40 GPU.
The server successfully initializes the new Cache Evictor with multi-queue routing, online parameter learning, and serves requests through the standard OpenAI-compatible API.
Under low-to-moderate concurrency with a single small model, vLLM's aggressive memory management rarely fills the KV cache to capacity, leading to limited eviction activity. This is consistent with prior observations that eviction policies have the largest impact under sustained cache pressure and heterogeneous workloads~\citep{DBLP:conf/usenix/GaoHSKJDYYZ24,DBLP:conf/iwqos/ChenZYHMYWW25}.
Therefore, for the following evaluation, we use trace-driven request injection under moderate-to-high load to induce sufficient cache pressure.

\subsection{Empirical Results}
\label{sec:empirical}
\begin{figure}[t]
    \centering
    \includegraphics[width=\textwidth]{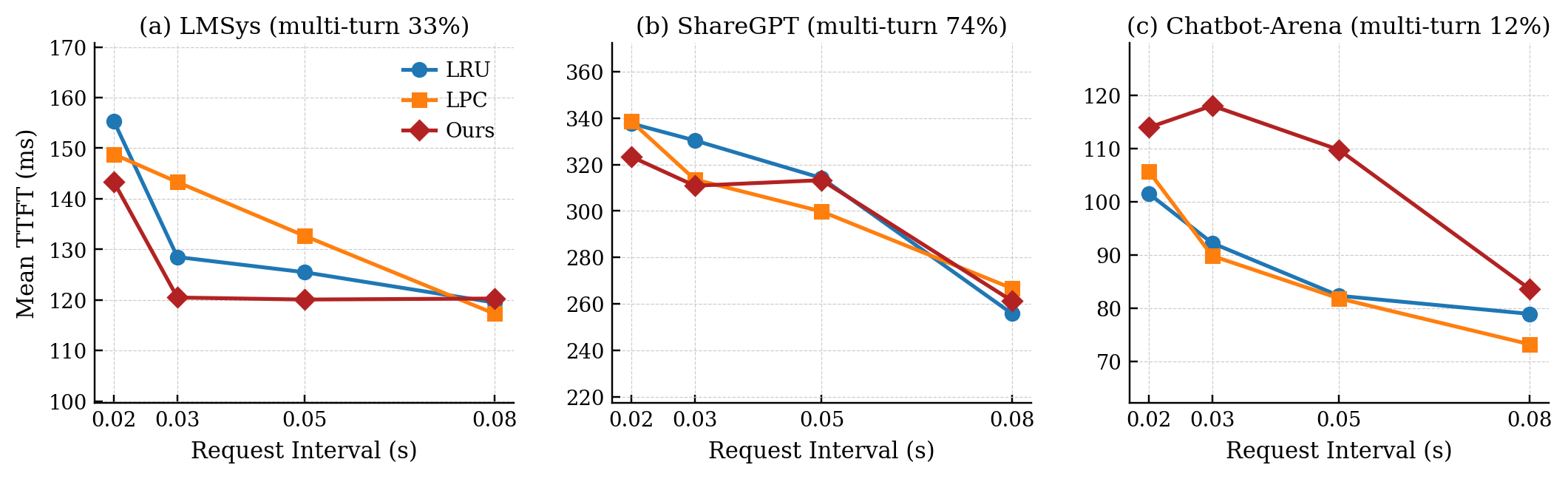}
    \caption{Mean TTFT vs.\ request interval across three conversational datasets, ordered by multi-turn ratio. SAECache achieves the lowest TTFT on LMSys ($33\%$ multi-turn) across all four interval settings, and matches the best baseline on ShareGPT ($74\%$ multi-turn) at high pressure (interval $\le 0.03$~s). On Chatbot-Arena ($12\%$ multi-turn), SAECache trails LPC and LRU; in this single-turn-dominated regime, the multi-queue management overhead exceeds the prefill savings from a higher hit rate (Figure~\ref{fig:hitrate}). The result is consistent with our design hypothesis: SAECache is most effective when reusable conversational prefixes are abundant.}
    \label{fig:ttft}
\end{figure}

\begin{figure}[t]
    \centering
    \includegraphics[width=\textwidth]{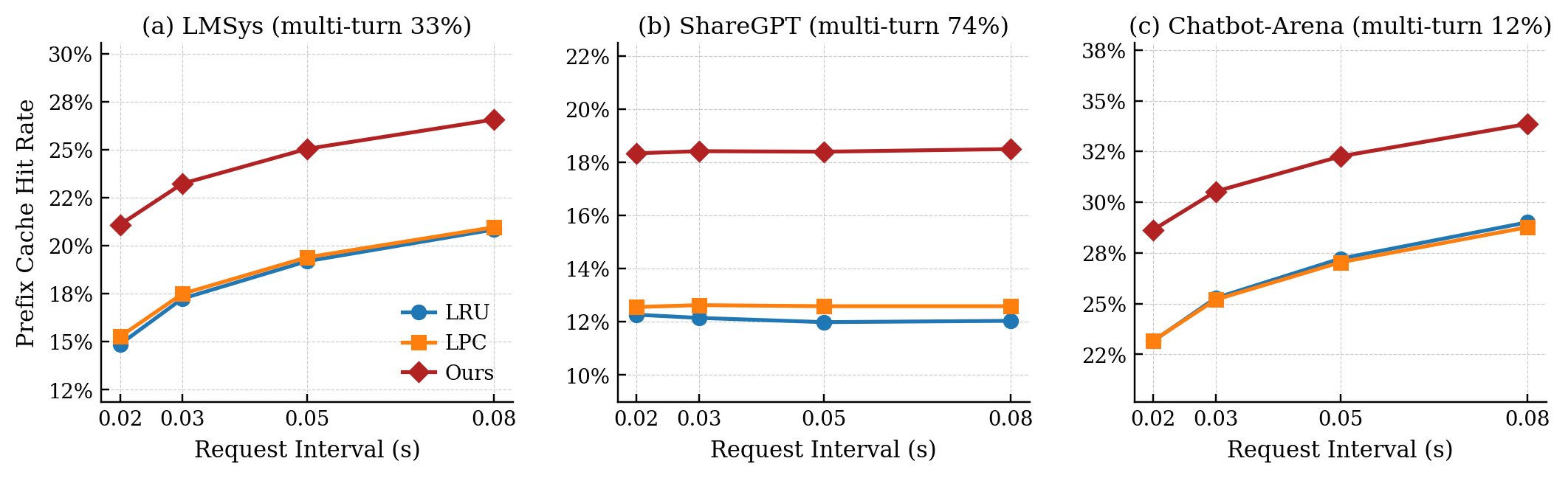}
    \caption{Prefix cache hit rate vs.\ request interval across three conversational datasets. SAECache achieves the highest hit rate in all $12$ (dataset, interval) cells, with a consistent margin of $4.8$--$5.9$ percentage points over both LRU and LPC. The gap is robust across multi-turn ratios spanning from $12\%$ (Chatbot-Arena) to $74\%$ (ShareGPT), confirming that the token-type-aware eviction policy correctly prioritises high-reuse blocks regardless of workload composition.}
    \label{fig:hitrate}
\end{figure}

We evaluate SAECache on three production-scale real datasets that span a wide range of multi-turn density: ShareGPT ($74\%$ multi-turn), LMSys ($33\%$ multi-turn), and Chatbot-Arena ($12\%$ multi-turn). 
For each dataset, we sweep the request injection interval over $\{0.02, 0.03, 0.05, 0.08\}$~s, all of which correspond to GPU-saturated and high-loaded regimes.
Figures~\ref{fig:ttft} and~\ref{fig:hitrate} report mean TTFT and prefix cache hit ratio for LRU, LPC, and SAECache.

\textbf{Cache hit ratio.} SAECache achieves the highest cache hit
rate in all $12$ $12$ dataset-interval configuration. Compared with the strongest baseline in each setting, SAECache improves the hit ratio by 
$4.8$--$5.9$ percentage points, as shown in Figure~\ref{fig:hitrate}.
The improvement is stable across the $6\times$ range of multi-turn density, suggesting
that our semantic-aware eviction policy consistently identifies high-reuse blocks
regardless of workload composition.

\textbf{TTFT latency.} 
The TTFT results in Figure~\ref{fig:ttft} show that higher cache hit ratio generally translates into lower prefill latency, but the magnitude depends on workload composition and system overhead. 
On LMSys, SAECache reduces mean TTFT by
$4$--$8\%$ compared with LRU and by up to $16\%$ compared with LPC. On ShareGPT, SAECache
matches the best baseline at highest pressure (i.e., inject interval $\le 0.03$~s), and remains
within $5\%$ at lower pressure. These results show that SAECache is particularly effective when the workload contains substantial multi-turn reuse.
On Chatbot-Arena, SAECache maintains the highest hit ratio but does not always achieve the lowest mean TTFT, trailing the
best baseline by $12$--$34\%$ depending on the injection interval. 
We attribute this regression to the overhead of multi-queue bookkeeping and multi-turn session predictor in Chatbot-Arena where only $12\%$ of requests belong to multi-turn sessions. In such workloads, the additional prefill savings from improved cache hits can be smaller than the overhead introduced by multi-queue management and the multi-turn session predictor. This suggests that our eviction policy consistently improves prefix reuse, but its latency gains become more visible when the workload has enough cache pressure and reusable prefixes to offset the extra policy overhead.

\textbf{Memory and computational overhead for predictor.}
Compared with LPC e5-small, SAECache achieves substantially higher inference efficiency while using significantly fewer resources. Specifically, SAECache reaches a throughput of 1{,}137 predictions per second, whereas LPC e5-small achieves only 129 predictions per second, making SAECache approximately $8.8\times$ faster. In terms of model size, SAECache contains only $1$M parameters, compared to $118$M parameters for LPC e5-small. This also results in a dramatic reduction in memory footprint: SAECache requires approximately $4$\,MB of memory, while LPC e5-small occupies around $472$\,MB.


\section{Conclusion}

This paper shows that prefix cache eviction in LLM serving should not treat all cached blocks uniformly, even when they have similar recency and frequency statistics. Through workload characterization, we find that prefix reuse depends strongly on token semantics, prompt structure, and session type, making existing policies insufficient for various heterogeneous workloads.
Motivated by these observations, we propose SAECache, a lightweight semantic-aware and adaptive eviction policy that separates different reuse mechanisms through a multi-queue design and learns token-type-aware eviction priorities from online cache feedback.
Our evaluation shows that SAECache consistently improves TTFT over all baselines across diverse workloads.

For future work, one important direction is to extend our semantic-aware eviction policy to the cluster-level setting, where request routing, prefix-cache locality, and per-node eviction decisions interact with each other.

\begin{algorithm}[t]
\caption{SAGECache Eviction}
\label{alg:global_eviction}
\begin{algorithmic}[1]
\Function{Evict}{}
    \State \textbf{// Stage 1: Drain evict-first queue}
    \If{$Q_{\mathrm{evict}} \neq \emptyset$}
        \State \Return $\Call{Remove}{\arg\min_{b \in Q_{\mathrm{evict}}} \mathrm{num\_tokens}(b)}$
    \EndIf

    \State \textbf{// Stage 2: Global score comparison across all queues}
    \State $b_{\mathrm{victim}} \gets \mathrm{nil}$
    \State $P_{\mathrm{best}} \gets +\infty$
    \For{each queue $q \in \{Q_{\mathrm{chat}}, Q_{\mathrm{agent}}, Q_{\mathrm{struct}}\}$}
        \For{each block $b \in q$}
            \State \textbf{// Eq.~(3): Global retention score}
            \State $P(b) \gets \alpha_q \cdot w_{\tau(b)} \cdot p_q(b) / \mathrm{ET}_b$
            \If{$P(b) < P_{\mathrm{best}}$}
                \State $P_{\mathrm{best}} \gets P(b)$
                \State $b_{\mathrm{victim}} \gets b$
            \EndIf
        \EndFor
    \EndFor

    \State \Return $\Call{Remove}{b_{\mathrm{victim}}}$
\EndFunction

\vspace{0.5em}

\Function{Add}{block $b$, cache hint $h$}
    \State \textbf{// Route block to appropriate queue}
    \State $q \gets \Call{Classify}{h}$
    \State $q.\mathrm{insert}(b)$

    \State \textbf{// Check miss-after-eviction for online learning}
    \If{$b.\mathrm{content\_hash} \in \mathrm{recently\_evicted}$}
        \State $\tau \gets \mathrm{recently\_evicted}[b.\mathrm{content\_hash}]$
        \State $\mathrm{token\_stats}[\tau].\mathrm{miss\_after\_evict} \mathrel{+}= 1$
    \EndIf

    \State \textbf{// Trigger parameter updates every $K$ evictions}
    \If{$\mathrm{eviction\_count} \bmod K = 0$}
        \State $\Call{TokenWeights}{}$
        \State $\Call{QueueWeights}{}$
        \State $\Call{LognormalParams}{}$
    \EndIf
\EndFunction

\vspace{0.5em}

\Function{Classify}{cache hint $h$}
    \If{$h.\mathrm{token\_type} \in \{\mathrm{cot}, \mathrm{decode}\}$ \textbf{or} $h$ is untemplated}
        \State \Return $Q_{\mathrm{evict}}$
    \EndIf
    \If{$h.\mathrm{is\_multi\_turn}$ \textbf{and} $h.\mathrm{is\_agentic}$}
        \State \Return $Q_{\mathrm{agent}}$
    \EndIf
    \If{$h.\mathrm{is\_multi\_turn}$ \textbf{or} $h.\mathrm{conversation\_id}$ is not empty}
        \State \Return $Q_{\mathrm{chat}}$
    \EndIf
    \If{$h.\mathrm{is\_shared\_prefix}$ \textbf{or} $h.\mathrm{token\_type} = \mathrm{system\_prompt}$}
        \State \Return $Q_{\mathrm{struct}}$
    \EndIf
    \State \Return $Q_{\mathrm{evict}}$
\EndFunction
\end{algorithmic}
\end{algorithm}








\begin{algorithm}[t]
\caption{Online Queue Weight}
\label{alg:queue_weight_update}
\begin{algorithmic}[1]
\Function{QueueWeights}{}
    \For{each queue $q \in \{Q_{\mathrm{chat}}, Q_{\mathrm{agent}}, Q_{\mathrm{struct}}\}$}
        \State $\mathrm{hits}_q \gets \mathrm{queue\_hits}[q]$
        \State $\mathrm{evictions}_q \gets \mathrm{queue\_evictions}[q]$
        \If{$\mathrm{evictions}_q > 5$}
            \State $\mathrm{efficiency}_q \gets \mathrm{hits}_q / \mathrm{evictions}_q$
            \State \textbf{// Target: efficiency normalized by temperature}
            \State $\mathrm{target} \gets 1.0 + \mathrm{efficiency}_q / T$
            \State \textbf{// EMA blend with previous weight}
            \State $\alpha_q \gets \alpha_q + \beta \cdot (\mathrm{target} - \alpha_q)$
            \State $\alpha_q \gets \mathrm{clamp}(\alpha_q, 0.1, 3.0)$
        \EndIf
        \State \textbf{// Reset counters for next window}
        \State $\mathrm{queue\_hits}[q] \gets 0$
        \State $\mathrm{queue\_evictions}[q] \gets 0$
    \EndFor
\EndFunction
\end{algorithmic}
\end{algorithm}

\bibliographystyle{plainnat}
\bibliography{references}


\appendix

\section{Related Work}
\label{app:related_work}
Block-based KV cache management has become a standard mechanism for accelerating LLM inference.
vLLM~\cite{DBLP:conf/sosp/KwonLZ0ZY0ZS23} and SGLang~\cite{DBLP:conf/nips/ZhengYXS0YCKSGB24} support prefix caching with production-style eviction mechanisms such as LRU and radix-tree-based management.
CachedAttention~\cite{DBLP:conf/usenix/GaoHSKJDYYZ24} extends prefix caching across memory tiers, while PREBLE~\cite{DBLP:conf/iclr/SrivatsaHAL025} studies cluster-level prefix sharing without changing per-node eviction.
Marconi~\cite{pan2025marconiprefixcachingera} optimizes prefix caching for hybrid attention architectures, and JENGA~\cite{DBLP:conf/sosp/ZhangDLKMWLYLLZ25} focuses on memory allocation for heterogeneous embeddings rather than fine-grained prefix eviction.

Recent work begins to incorporate workload structure into KV cache management. For example, CONTINUUM~\cite{DBLP:journals/corr/abs-2511-02230} introduces time-to-live-based KV pinning for agentic tool cal, and KVFlow~\cite{pan2025kvflowefficientprefixcaching} prioritizes blocks according to agent execution order.
These designs improve KV reuse in specific agentic scenarios but do not generalize to arbitrary workloads. 

The closest work to ours is LPC~\cite{yang2026learned}, which trains a 118M-parameter text embedding model to predict conversation continuation probability and combines it with a manually tuned time-decay rule.
However, LPC assigns a uniform conversation-level score to all blocks in the same session, so semantically different blocks, such as system prompts and model responses, receive the same eviction priority.
SAECache differs from LPC in three ways.
First, it makes eviction decisions at token-level granularity: each block receives its own retention score based on token type and position, rather than inheriting a uniform conversation-level score.
Second, SAECache does not require an external embedding model or periodic retraining. Its multi-turn session predictor directly reuses hidden states already produced during serving, while key parameters are updated online. This design reduces serving overhead and improves robustness under workload shifts.
Third, SAECache decouples heterogeneous reuse patterns through a multi-queue design: multi-turn queues capture session-local temporal reuse, whereas the structural queue captures shared-template reuse in single-turn prompts. In contrast, the single continuation probability used by LPC collapses these distinct reuse mechanisms into one scalar and therefore cannot express their structural differences.

\section{Adaptive Parameter Learning}\label{adaptive}
In this section, we provide the details for our adaptive parameter update rule.

\begin{figure}
    \centering
    \includegraphics[width=0.8\linewidth]{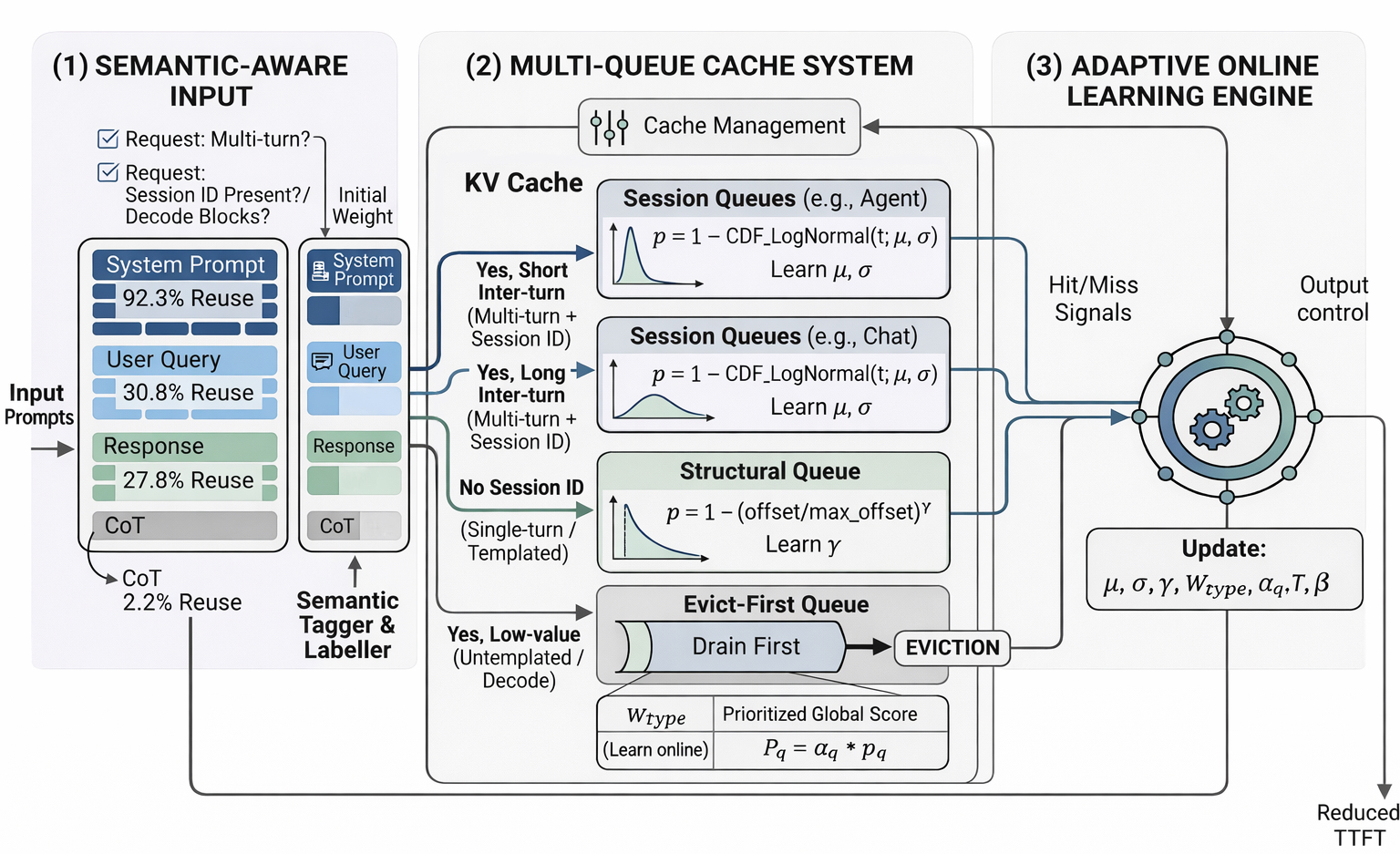}
    \caption{Overview of the SAECache eviction workflow. }
    \label{fig:framework_overview}
\end{figure}

\textbf{Token-type weight update.}
Token-type weights are updated using an exponential moving average (EMA) based on observed hit and miss feedback:
\[
w_{\text{target}} = 1.0 + r_{\text{miss}} \times 5.0 + r_{\text{reuse}} \times 2.0,
\]
and
\[
w_{\text{new}} = (1 - \eta) \times w_{\text{old}} + \eta \times w_{\text{target}},
\]
where $r_{\text{miss}}$ is the miss-after-eviction rate, i.e., the fraction of evicted blocks of this type that are later requested again, $r_{\text{reuse}}$ is the overall cache hit rate for this type, and $\eta$ is the learning rate.
The miss-after-eviction rate provides direct feedback on eviction quality: if blocks of a given type are frequently evicted and then requested again, their weight increases.
The reuse rate captures the intrinsic reuse value of each token type.
To handle non-stationary workloads, all statistics are maintained in a sliding window with exponential decay factor $0.99$.

\begin{algorithm}[!ht]
\caption{Online Token Weight}
\label{alg:token_weight_update}
\begin{algorithmic}[1]
\Function{TokenWeights}{}
    \For{each token type $\tau \in \{\mathrm{system\_prompt}, \mathrm{user\_query}, \mathrm{tool\_output}, \mathrm{response}, \mathrm{cot}\}$}
        \State $\mathrm{evicted} \gets \mathrm{token\_stats}[\tau].\mathrm{evicted}$
        \State $\mathrm{missed} \gets \mathrm{token\_stats}[\tau].\mathrm{miss\_after\_evict}$
        \If{$\mathrm{evicted} > 10$}
            \State $\mathrm{miss\_rate} \gets \mathrm{missed} / \mathrm{evicted}$
            \State \textbf{// High miss rate means this type was evicted too aggressively}
            \State $w_{\tau} \gets w_{\tau} \cdot (1 + \eta \cdot \mathrm{miss\_rate})$
            \State $w_{\tau} \gets \mathrm{clamp}(w_{\tau}, 0.1, 5.0)$
        \EndIf
        \State \textbf{// Decay statistics for temporal locality}
        \State $\mathrm{token\_stats}[\tau].\mathrm{evicted} \gets \lfloor \lambda \cdot \mathrm{evicted} \rfloor$
        \State $\mathrm{token\_stats}[\tau].\mathrm{miss\_after\_evict} \gets \lfloor \lambda \cdot \mathrm{missed} \rfloor$
    \EndFor
\EndFunction
\end{algorithmic}
\end{algorithm}

\begin{figure}[!ht]
    \centering
    \includegraphics[width=0.7\linewidth]{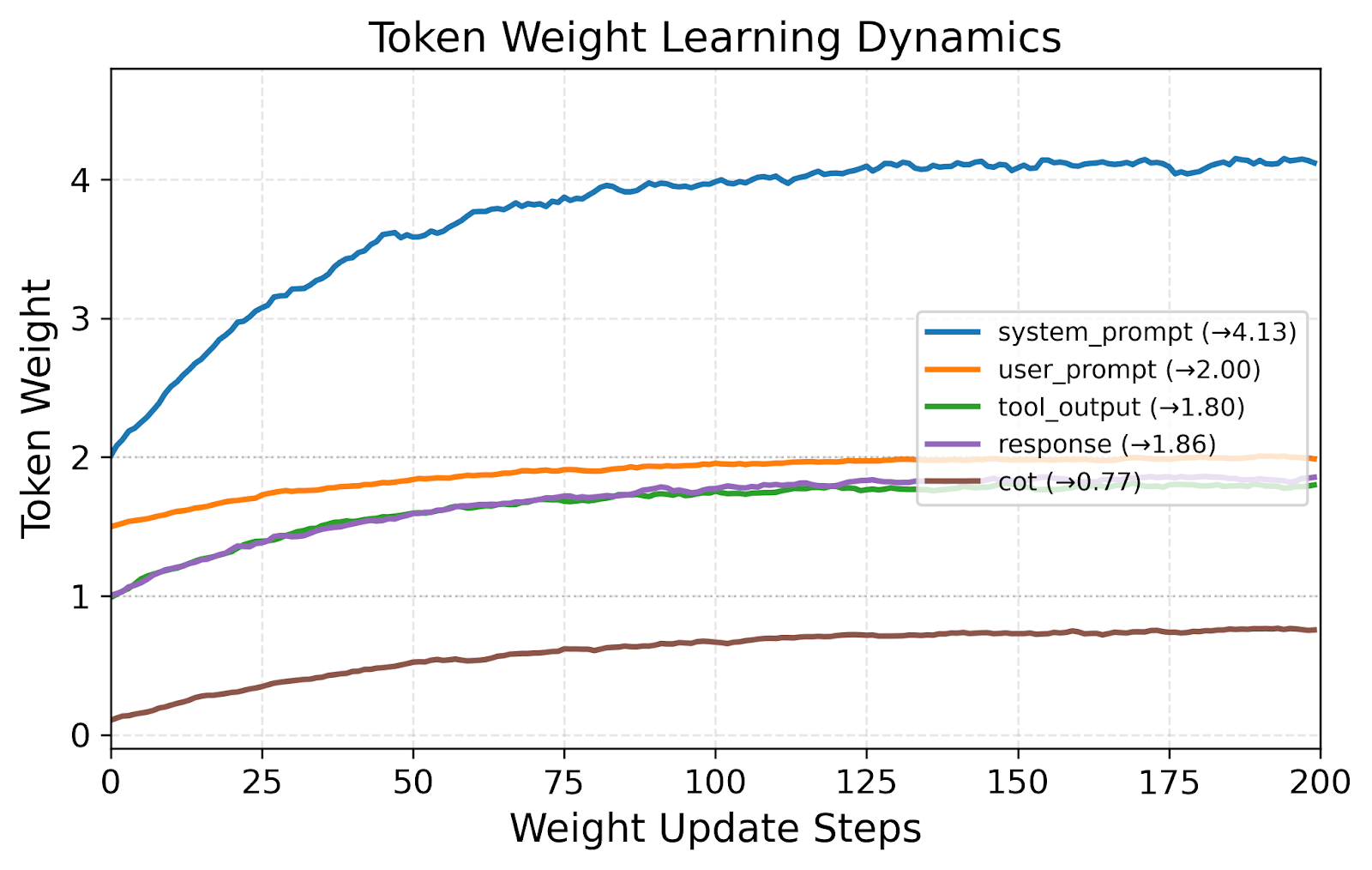}
    \caption{Token weight learning dynamics.  The system prompt weight increases from an initial value of 2.0 to 4.13, corresponding to a 107\% increase, which reflects its 92.3\% reuse rate. The user prompt weight rises moderately, while the chain-of-thought weight remains the lowest, which matches its 2.2\% reuse rate. }
    \label{fig:token_weight_learning}
\end{figure}

We validate this update rule on the \texttt{balanced} workload, in which all five token types receive sufficient exposure for the online statistics to converge. As shown in Figure~\ref{fig:token_weight_learning}, starting from a uniform initialization, the learner converges within roughly $200$ update steps. The converged weights satisfy
\begin{equation}
    w_{\text{sys}} > w_{\text{usr}} > w_{\text{resp}} \approx w_{\text{tool}} > w_{\text{cot}},
    \label{eq:learned-weight-ordering}
\end{equation}
with values $4.13$, $2.00$, $1.86$, $1.80$, and $0.77$ respectively. This ordering closely tracks the empirical reuse-rate ordering reported in Section~\ref{sec:2.3tokentypes}. In particular, the system-prompt weight rises by $107\%$ from its initial value of $2.0$, which is the largest gain among all types and is consistent with system prompts having the highest measured reuse rate of $92.3\%$. No reuse-rate ordering or weight magnitudes are supplied to the learner; both emerge from eviction feedback alone.
The pseudo code for updating can be found in Alg.\ref{alg:token_weight_update}.

\textbf{Log-normal parameter updates.} 
The log-normal parameters $\mu$ and $\sigma$ that govern multi-turn session reuse timing vary significantly across multi-turn session types and may not match pre-fitted values when workload changes. We use online maximum likelihood estimation: when cache hits occur for multi-turn blocks, we record the elapsed time since last access and periodically compute:
\[
\mu_{\text{MLE}} = \mathrm{mean}(\ln(\bm{t})), \quad \sigma_{\text{MLE}} = \mathrm{std}(\ln(\bm{t})).
\]
where $\mathbf{t}$ is the vector of observed inter-access times. Estimates are smoothed via EMA:
\[
\mu_{\text{new}} = \beta \times \mu_{\text{old}} + (1 - \beta) \times \mu_{\text{MLE}}.
\]
Separate estimators are maintained for each multi-turn session type (chat vs.\ agentic), with a minimum sample threshold ($n \geq 10$) before updates begin. The EMA factor $\beta$ itself adapts: high observation variance triggers lower $\beta$ for faster tracking, while low variance triggers higher $\beta$ for stability.

\begin{algorithm}[!ht]
\caption{Online Log-Normal Parameter}
\label{alg:lognormal_update}
\begin{algorithmic}[1]
\Function{LognormalParams}{}
    \For{each style $s \in \{\mathrm{chat}, \mathrm{agent}\}$}
        \State $\mathrm{intervals} \gets \mathrm{reuse\_intervals}[s]$
        \If{$|\mathrm{intervals}| > 20$}
            \State \textbf{// Online MLE for log-normal}
            \State $\mathrm{log\_intervals} \gets [\ln(\Delta t) \; \text{for each } \Delta t \in \mathrm{intervals}]$
            \State $\mu_{\mathrm{sample}} \gets \mathrm{mean}(\mathrm{log\_intervals})$
            \State $\sigma_{\mathrm{sample}} \gets \mathrm{std}(\mathrm{log\_intervals})$
            \State \textbf{// EMA blend with current parameters}
            \State $\mu_s \gets \mu_s + \beta_{\mathrm{ln}} \cdot (\mu_{\mathrm{sample}} - \mu_s)$
            \State $\sigma_s \gets \sigma_s + \beta_{\mathrm{ln}} \cdot (\sigma_{\mathrm{sample}} - \sigma_s)$
            \State $\sigma_s \gets \max(\sigma_s, 0.1)$
            \State \textbf{// Truncate buffer to bound memory}
            \State $\mathrm{reuse\_intervals}[s] \gets \mathrm{intervals}[-200:]$
        \EndIf
    \EndFor
\EndFunction
\end{algorithmic}
\end{algorithm}

\textbf{Decay power update.}
The decay power, e.g., $\gamma$ in \eqref{position_rule}, used in the structural queue's local priority rule is updated online from observed position-hit correlations.
Recall that the structural-queue priority is defined as

\[
p = 1 - \left(\frac{\mathrm{offset}}{\mathrm{max\_offset}}\right)^{\mathrm{decay\_power}}.
\]
We divide the prompt into positional bins, track hit rates per bin, and compare front-half versus back-half hit rates, where
\[
\mathrm{ratio} = \frac{\mathrm{back\_avg\_hit\_rate}}{\mathrm{front\_avg\_hit\_rate}},
\]
and from this we derive
\[
\mathrm{estimated\_power} = \frac{1.0}{\mathrm{ratio} + 0.1}.
\]
A smaller ratio indicates weaker reuse in later prompt positions and therefore leads to a larger decay power, which more aggressively deprioritizes later blocks.
The estimated power is smoothed with an exponential moving average and clipped to the range $[0.3, 3.0]$.
In our experiments, the learned power ranges from 0.94 to 1.0 depending on the workload composition.

We note that position and token type are correlated in typical prompt structures, because system prompts occupy early positions while user queries appear later. However, position-based priority remains valuable as a complementary signal, because it provides intra-type differentiation for long homogeneous segments and serves as a useful prior when token-type annotation is unavailable or ambiguous. Our ablation study shows that position decay learning contributes an additional 12\% improvement on top of token-type weighting.

\begin{figure}
    \centering
    \includegraphics[width=0.8\linewidth]{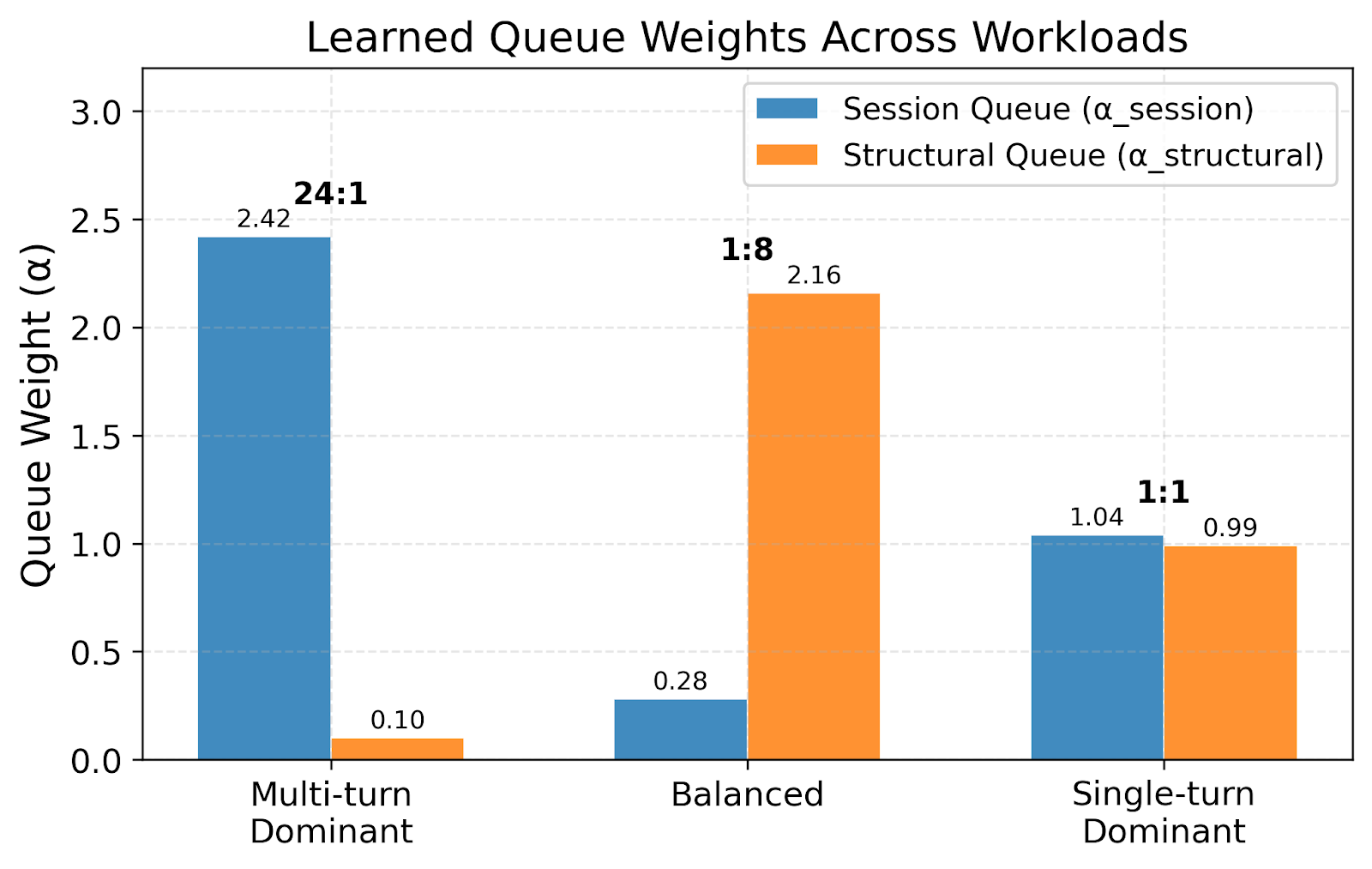}
    \caption{Learned queue weights across workloads. This Figure shows the converged queue weights under different workloads, demonstrating that SAECache dynamically adapts its cache allocation strategy to diverse workload compositions. Under the multi-turn-dominant workload, the multi-turn queue weight increases to $2.42$ while the structural queue weight decreases to $0.10$, resulting in an approximate ratio of $24{:}1$, which strongly prioritizes conversational reuse. In the balanced workload, the ratio reverses to approximately $1{:}8$, while under the single-turn-dominant workload, both weights converge to values close to $1.0$, indicating balanced cache treatment. These results demonstrate that SAECache can automatically redistribute cache priority across queues according to workload characteristics without requiring manual parameter tuning.}
    \label{fig:queue_weights}
\end{figure}

\textbf{Queue weight update.}
For each queue $q$ except the evict-first queue, we compute its recent hit efficiency as
\[
E_q = \frac{H_q}{C_q},
\]
where $H_q$ denotes the number of hit tokens contributed by queue $q$ over a recent window, and $C_q$ denotes its capacity fraction.
Thus, $E_q$ measures the cache benefit obtained per unit of allocated capacity.
We update the queue weight using temperature-scaled relative efficiency:
\[
\alpha_q = \beta \times \alpha_q + (1 - \beta) \times \left(\frac{E_q}{\bar{E}}\right)^{1/T},
\]
where $\bar{E}$ is the mean efficiency across queues and $T$ is a temperature parameter that controls how strongly the policy differentiates between high- and low-efficiency queues. A lower temperature amplifies efficiency differences and leads to more aggressive reweighting, while a higher temperature produces smoother and more conservative updates. 
Figure~\ref{fig:queue_weights} shows the converged queue weights under different workloads, demonstrating that our update rule automatically adapts to diverse workload composition.
In the multi-turn-dominant workload, the multi-turn queue weight reaches 2.42 while the structural weight is 0.10, corresponding to a ratio of 24:1.
In the balanced workload, the ratio reverses to approximately 1:8, while in the single-turn-dominant workload, the two weights converge to around 1.0.
These results show that SAECache can automatically reallocate cache priority across queues without manual tuning.
The updating algorithm can be found in Alg.\ref{alg:queue_weight_update}.


\section{Sensitivity Analysis of Hyperparameters}
\label{app:sensitivity}

\paragraph{Sensitivity to $\alpha$ and $\beta$.}
We evaluate the token weight update rule
\[
w_{\text{target}} = 1.0 + \alpha \cdot \text{miss\_rate} + \beta \cdot \text{reuse\_rate}
\]
across a grid of 30 $(\alpha, \beta)$ configurations with 
$\alpha \in \{0.5, 1, 2, 5, 10, 20\}$ and 
$\beta \in \{0.5, 1, 2, 5, 10\}$ 
on the balanced workload.

Across all configurations, SAECache achieves the same TTFT of 3.564s and cache hit ratio of 18.9\%, showing that its performance is insensitive to these hyperparameters.
The learned weights converge to similar relative orderings for all choices of $\alpha$ and $\beta$: \textit{system\_prompt} consistently receives the highest weight, while \textit{chain-of-thought} tokens remain the lowest-weight type.
This suggests that the cache-efficiency gain comes mainly from learning the correct relative ranking among token types, rather than from carefully tuning the specific update coefficients.

\section{Ablation Study}
\label{ablation_study}

\begin{figure}
    \centering
    \includegraphics[width=0.7\linewidth]{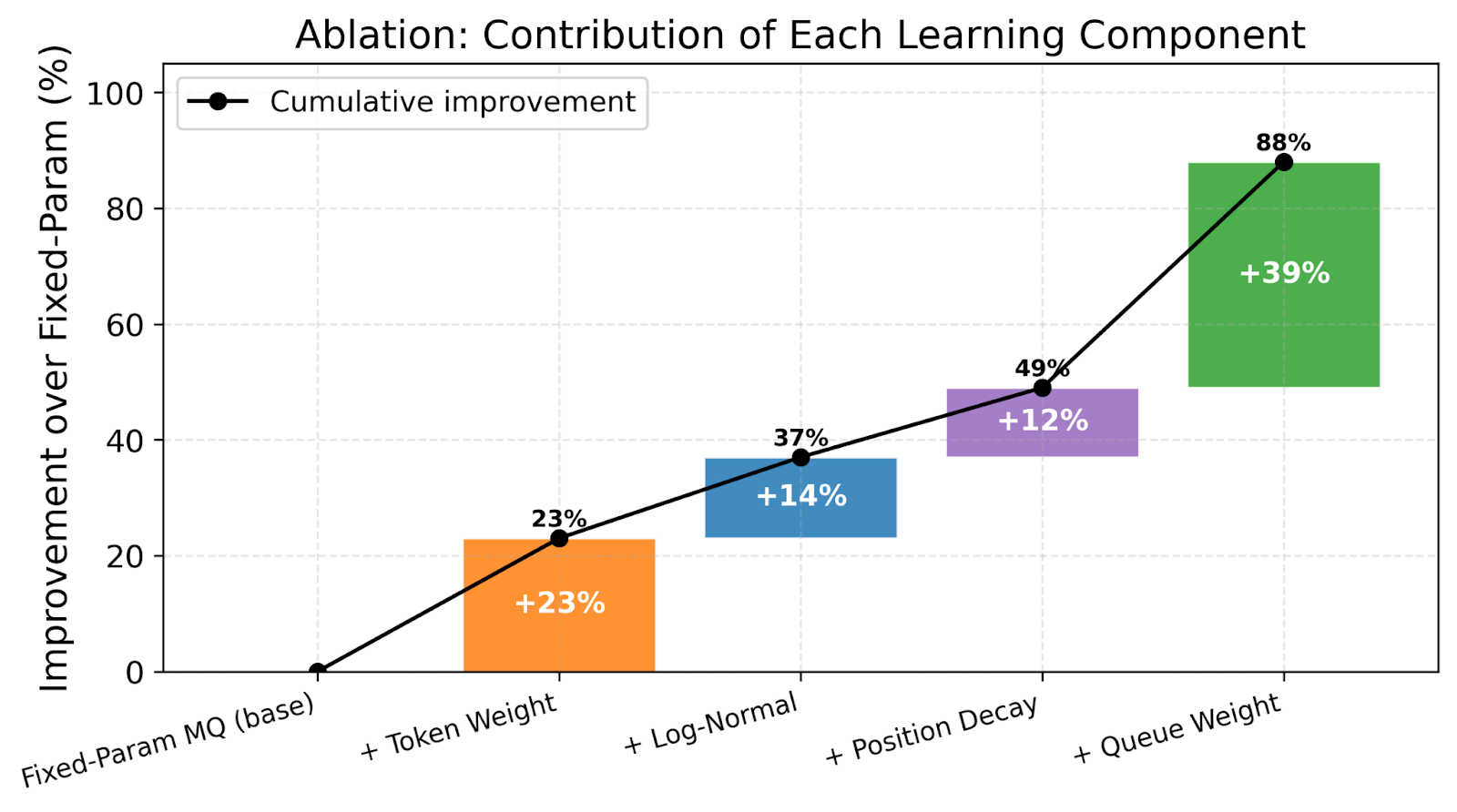}
    \caption{Ablation results. Each bar incrementally adds one learning component to the Fixed-Param Multi-Queue baseline. Queue-weight learning and token-type weight learning are the two largest contributors.}
    \label{fig:ablation_results}
\end{figure}

In this section, we further examine how each learning component contributes to the overall performance of SAECache.
Starting from the Fixed-Param (Multi-Queue) baseline, we incrementally add token-type weight learning, log-normal timing learning, position-decay learning, and queue-weight learning.
Figure~\ref{fig:ablation_results} reports the cumulative improvement from these components.
We also compare the variants of our method as follows: (a) Token-Weight-Only, which uses the semantic-aware token weighting from Section~3.2 without the multi-queue architecture to isolate the contribution of token-type awareness; and (b) Fixed-Param Multi-Queue, which uses the full multi-queue architecture with parameters fitted offline from chat-dominated traces, thereby isolating the benefit of online adaptation.

\paragraph{Synthetic Workload Composition}
\label{sec:synthetic-workloads}

We use four synthetic workloads generated by our request simulator, in which prompt length, output length, and prefix-sharing ratio are sampled from preset distributions rather than drawn from real conversational logs such as ShareGPT or LMSys. Synthetic generation is necessary here because no single real dataset spans the full range of multi-turn ratios required to isolate each learning component's contribution; results on real datasets appear in Section~\ref{sec:empirical} and Figures~\ref{fig:ttft}--\ref{fig:hitrate}.

The first workload, \texttt{tool\_use}, is a homogeneous single-turn workload with $85\%$ shared prefix and $282$ requests, used as a high-contention microbenchmark. The remaining three workloads vary the multi-turn / single-turn composition by mixing five request categories, summarised in Table~\ref{tab:workload-composition}: Chat (multi-turn conversation), Agent (agentic interaction), Tool Use (function-calling requests), Programming (code-related requests), and Doc QA (document question answering).

\begin{table}[h]
\centering
\caption{Per-category proportions for the three multi-turn-mixture workloads. The \texttt{tool\_use} microbenchmark is a separate single-category workload and is not included here.}
\label{tab:workload-composition}
\small
\begin{tabular}{lccccc|cc}
\toprule
Workload & Chat & Agent & Tool Use & Programming & Doc QA & Multi-turn & Single-turn \\
\midrule
\texttt{multi\_turn\_dominant}  & 50\% & 30\% & 10\% & 5\%  & 5\%  & 80\% & 20\% \\
\texttt{balanced}               & 30\% & 20\% & 25\% & 15\% & 10\% & 50\% & 50\% \\
\texttt{single\_turn\_dominant} & 10\% & 10\% & 40\% & 25\% & 15\% & 20\% & 80\% \\
\bottomrule
\end{tabular}
\end{table}

\paragraph{multi\_turn\_dominant.} Multi-turn categories (Chat + Agent) account for $80\%$ of all traffic, with Tool Use, Programming, and Doc QA together making up the remaining $20\%$. This mixture models conversational deployments with high context reuse, such as customer support assistants or long-horizon agent sessions.

\paragraph{balanced.} The five categories are distributed roughly evenly, with multi-turn and single-turn traffic each accounting for $50\%$. This mixture serves as a neutral baseline for evaluating scheduler behaviour in the absence of a dominant traffic mode.

\paragraph{single\_turn\_dominant.} Single-turn categories (Tool Use + Programming + Doc QA) account for $80\%$ of all traffic; multi-turn traffic is compressed to $20\%$. This mixture models deployments dominated by one-shot requests, such as code completion services or standalone function-calling APIs.

Together, the three multi-turn-mixture workloads span the full spectrum from multi-turn-dominated to single-turn-dominated traffic, allowing us to evaluate whether the scheduler maintains stable cache hit rate and TTFT under \emph{workload mismatch}---i.e., when the request composition deviates substantially from the trace used to fit any offline parameters.

\textbf{Component-wise contribution.} Token-type weight learning improves performance by 23\%, confirming the effectiveness of our semantic-aware token weighting in eviction decisions. Log-normal timing learning and position-decay learning provide additional gains of 14\% and 12\%, respectively.
These gains show that although the default timing and position parameters provide reasonable initial estimates, adapting them online still improves cache decisions.
Queue-weight learning contributes the largest improvement, adding 39\% over the preceding configuration.
This highlights the importance of dynamically reallocating cache priority across queues, especially when the workload composition shifts between multi-turn and single-turn traffic. In summary, the four learning components together yield an 88\% improvement over the Fixed-Param baseline.

\textbf{Learned token-type weights.}
We next examine whether the learned parameters match the empirical reuse patterns observed in Section~\ref{sec:prefixreuse} (shown in Figure~\ref{fig:token_weight_learning}). Starting from their initial values, the token-type weights converge to an ordering that closely follows the measured reuse hierarchy.
The system-prompt weight increases from 2.0 to 4.13, corresponding to a 107\% increase and the largest absolute gain, consistent with its high reuse rate of 92.3\%.
The user-prompt weight increases from 1.5 to 2.00, while the tool-output and response weights increase from 1.0 to 1.80 and 1.86, respectively.
The chain-of-thought weight increases from 0.1 to 0.77, correcting an overly aggressive initial value, but remains the lowest-weight type, consistent with its low reuse rate of 2.2\%.
Overall, the learned ordering,
\[
\textit{system prompt} > \textit{user prompt} > \textit{response} \approx \textit{tool output} > \textit{chain-of-thought},
\]
matches the empirical reuse ordering, showing that SAECache can discover the relative value of token types from cache feedback without manual specification.


\textbf{Learned queue weights across workloads.} Queue weights also adapt to workload composition.
On \texttt{multi\_turn\_dominant}, SAECache learns a multi-turn queue weight of 2.42 and a structural queue weight of 0.10, corresponding to a 24:1 ratio that correctly prioritizes multi-turn session reuse.
On \texttt{single\_turn\_dominant}, the ratio reverses: the multi-turn queue weight becomes 0.28 and the structural queue weight becomes 2.16, approximately a 1:8 ratio.
These results show that SAECache can automatically shift priority toward the queue type that provides the highest reuse benefit under each workload, without workload-specific tuning.

\section{Log-Normal Modeling of Inter-Turn Intervals}
\label{app:lognormal-params}

This section validates the log-normal modeling assumption used 
used in our policy for 
within-session inter-turn intervals, and analyzes the parameters
recovered by our online learning algorithm.

\textbf{Datasets and pre-processing.}
We use two representative multi-turn traces. \emph{Qwen-Bailian}~\cite{qwenbailian} is a two-hour anonymized production chat trace from the
Qwen serving cluster on Aliyun Bailian. We reconstruct sessions by following
the \texttt{parent\_chat\_id} chains in the JSONL records and compute the
inter-turn interval as the difference between a child request's timestamp and
its parent's. \emph{CC-Bench-V1.1}~\cite{ccbench} contains 370 complete
Claude-Code agentic trajectories spanning 5 LLMs and 6 task categories, and we
treat each user-side message (including tool-result auto-fills) as one turn.
For both traces, we discard intervals $\leq 0$ (logging artifacts) and
$\geq 24$h (treated as session boundaries). For CC-Bench, we additionally
apply a 0.1\,s minimum-interval filter to remove sub-millisecond logging
artifacts at the SDK precision floor (193 intervals at exactly 0.001\,s, 621
intervals below 0.1\,s), and we verify robustness to this threshold below
(Table~\ref{tab:thresh-sweep}). After processing we obtain
$N_\text{chat} = 19{,}957$ intervals across 9{,}012 sessions in Bailian, and
$N_\text{agentic} = 15{,}755$ intervals across 370 sessions in CC-Bench.

\begin{figure}[!ht]
\centering
\includegraphics[width=\linewidth]{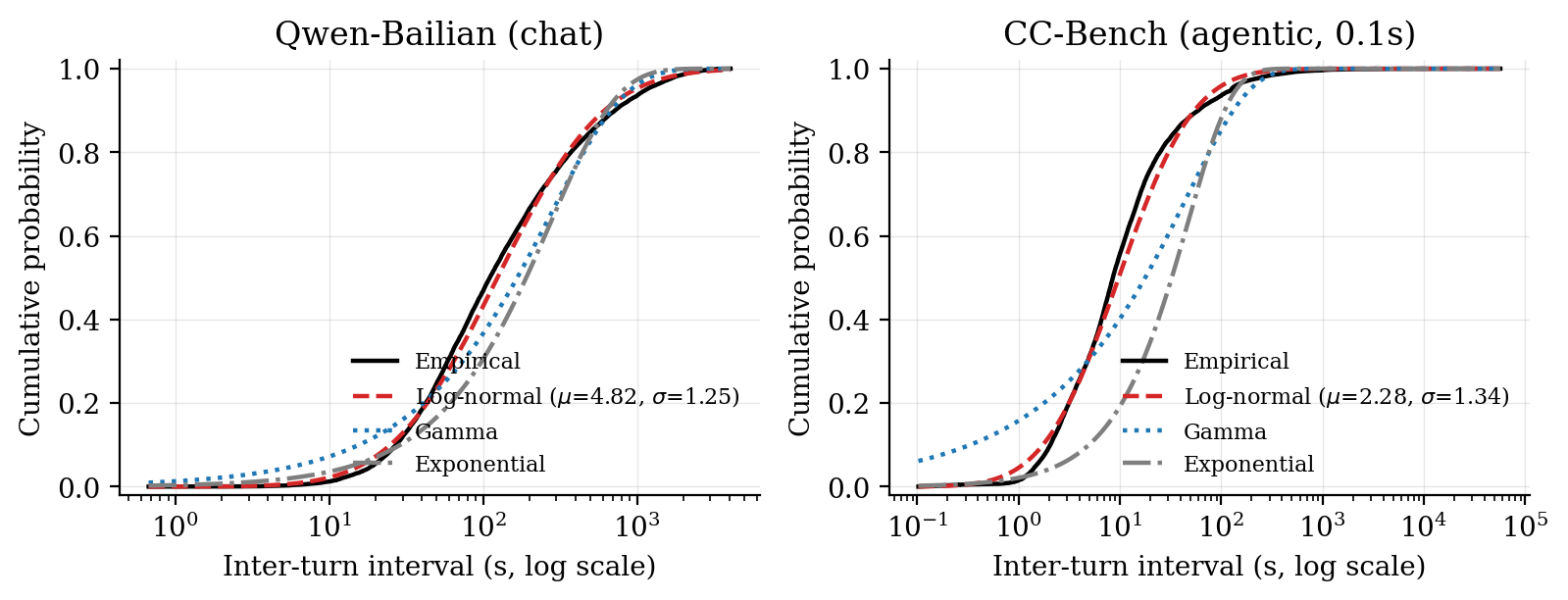}
\caption{Empirical CDF (black) overlaid with fitted log-normal (red dashed),
Gamma (blue dotted), and Exponential (grey dash-dot) CDFs.
Log-normal closely tracks the empirical distribution across the full
support on both traces.}
\label{fig:lognormal-cdf}
\end{figure}

\begin{figure}[!ht]
\centering
\includegraphics[width=\linewidth]{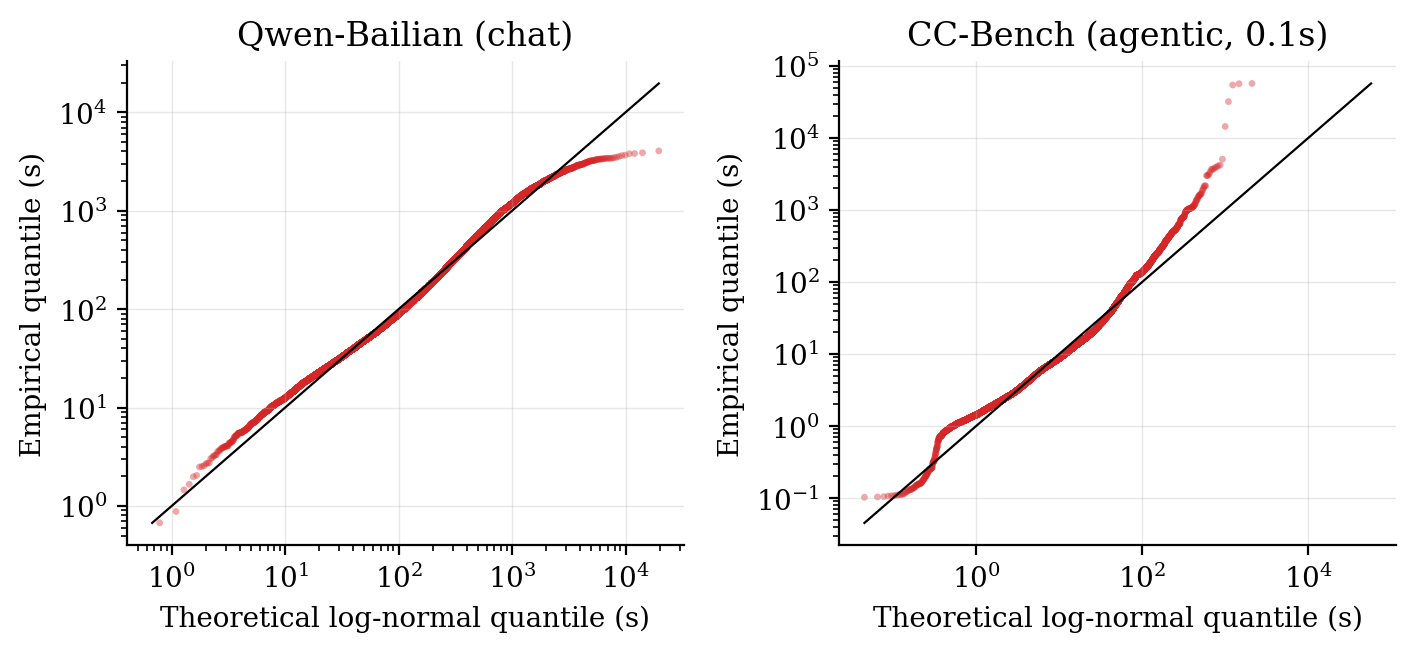}
\caption{Q--Q plot of empirical inter-turn intervals against the fitted
log-normal quantiles. Points lie close to the diagonal across the bulk of
the distribution; mild deviation in the extreme upper tail
(Bailian above P99.5, CC-Bench above P95) reflects somewhat heavier
empirical tails than log-normal.}
\label{fig:lognormal-qq}
\end{figure}

\textbf{Empirical fit.} For each trace we fit a log-normal distribution by maximum likelihood
estimation on $\log \Delta t$, and compare against Exponential and Gamma
baselines. Goodness-of-fit is reported via the Kolmogorov--Smirnov (K-S)
statistic $D$ and the coefficient of determination $R^2$ between empirical
and fitted CDFs. Table~\ref{tab:workload-summary} summarizes the per-trace
characteristics, and Table~\ref{tab:lognormal-fits} reports the cross-distribution
comparison. On both traces, log-normal achieves the smallest K-S statistic
($D \leq 0.060$) and the highest $R^2$ ($\geq 0.987$) by a wide margin;
Gamma is consistently second-best with 2--4$\times$ larger K-S distance, and
Exponential is strongly rejected. Figure~\ref{fig:lognormal-cdf} overlays the
empirical CDF with the three fitted CDFs on a logarithmic time axis, and
Figure~\ref{fig:lognormal-qq} provides the corresponding Q--Q plots against
the fitted log-normal.

\begin{table}[!ht]
\caption{Inter-turn interval characteristics: chat vs.\ agentic multi-turn workloads.}
\label{tab:workload-summary}
\centering
\small
\setlength{\tabcolsep}{12pt}
\renewcommand{\arraystretch}{1.18}
\begin{tabular}{@{}l S[table-format=5.1] S[table-format=5.1]@{}}
\toprule
                                & {\textbf{Bailian}}  & {\textbf{CC-Bench}}  \\
                                & {\textbf{(chat)}}   & {\textbf{(agentic)}} \\
\midrule
\# intervals                    & 19957 & 15755 \\
\# sessions                     &  9012 &   370 \\[2pt]
\multicolumn{3}{@{}l}{\textit{Empirical percentiles (seconds)}} \\
\quad P50                       &  110.6 &    8.5 \\
\quad P80                       &  372.4 &   25.0 \\
\quad P95                       & 1151.9 &  125.8 \\
\quad P99                       & 2207.3 &  453.6 \\
\quad P99 / P50 ratio           &   19.9 &   53.0 \\[2pt]
\multicolumn{3}{@{}l}{\textit{Fitted log-normal parameters}} \\
\quad $\mu$ (log-scale)         &    4.82 &    2.28 \\
\quad $\sigma$ (log-scale)      &    1.25 &    1.34 \\[2pt]
\multicolumn{3}{@{}l}{\textit{Goodness-of-fit (log-normal vs.\ alternatives)}} \\
\quad K-S statistic $D$         &  \bfseries 0.038 & \bfseries 0.060 \\
\quad $R^2$                     &  \bfseries 0.994 & \bfseries 0.987 \\
\bottomrule
\end{tabular}
\end{table}

\begin{table}[!ht]
\caption{Goodness-of-fit comparison across candidate distributions.
Log-normal achieves the smallest K-S statistic and the highest $R^2$
on both traces by a wide margin.}
\label{tab:lognormal-fits}
\centering
\small
\setlength{\tabcolsep}{8pt}
\renewcommand{\arraystretch}{1.15}
\begin{tabular}{@{}llrlrrr@{}}
\toprule
Trace & Distribution & $N$ & Parameters & K-S $D$ & $R^2$ & AIC \\
\midrule
\multirow{3}{*}{Bailian (chat)}
  & \textbf{Log-normal} & 19{,}957 & $\mu{=}4.82, \sigma{=}1.25$
      & \textbf{0.038} & \textbf{0.994} & \textbf{2.58e+5} \\
  & Gamma       & 19{,}957 & shape=0.75, scale=366
      & 0.118 & 0.937 & 2.63e+5 \\
  & Exponential & 19{,}957 & rate=0.0036
      & 0.172 & 0.864 & 2.64e+5 \\
\midrule
\multirow{3}{*}{CC-Bench (agentic)}
  & \textbf{Log-normal} & 15{,}755 & $\mu{=}2.28, \sigma{=}1.34$
      & \textbf{0.060} & \textbf{0.987} & \textbf{1.26e+5} \\
  & Gamma       & 15{,}755 & shape=0.42, scale=113
      & 0.238 & 0.757 & 1.41e+5 \\
  & Exponential & 15{,}755 & rate=0.0213
      & 0.418 & 0.098 & 1.50e+5 \\
\bottomrule
\end{tabular}
\end{table}

\textbf{Sensitivity to the CC-Bench filtering threshold.} To verify that our findings are not an artifact of the specific 0.1\,s
threshold, we sweep the threshold from 0\,s to 3\,s
(Table~\ref{tab:thresh-sweep}). Fit quality is stable across the
0.1--1.0\,s regime ($R^2 \geq 0.984$, K-S $\leq 0.064$). Without filtering,
the K-S statistic doubles to 0.118 because timestamp-precision artifacts at
$\sim 10^{-3}$\,s create a degenerate second mode that no single-component
log-normal can fit. With overly aggressive filtering ($\geq 2.0$\,s), fit
quality also degrades because legitimate short agent steps are discarded.

\begin{table}[!ht]
\caption{Sensitivity of the CC-Bench log-normal fit to the minimum-interval
threshold. The 0.1\,s row (bold) is the operating point used in this work.}
\label{tab:thresh-sweep}
\centering
\small
\setlength{\tabcolsep}{9pt}
\renewcommand{\arraystretch}{1.18}
\begin{tabular}{@{}S[table-format=1.1] S[table-format=5.0]
                  S[table-format=3.1] S[table-format=2.2] S[table-format=2.2]
                  S[table-format=1.3] S[table-format=1.3]
                  S[table-format=1.3] S[table-format=1.3]@{}}
\toprule
{Threshold (s)} & {$N$} & {\% kept}
      & {P50 (s)} & {P80 (s)}
      & {$\mu$} & {$\sigma$}
      & {K-S $D$} & {$R^2$} \\
\midrule
0.0 & 16376 & 100.0 &  8.13 & 23.65 & 2.010 & 1.911 & 0.118 & 0.945 \\
\bfseries 0.1 & \bfseries 15755 & \bfseries 96.2 & \bfseries 8.54 & \bfseries 25.01 & \bfseries 2.278 & \bfseries 1.343 & \bfseries 0.060 & \bfseries 0.987 \\
0.5 & 15659 &  95.6 &  8.61 & 25.19 & 2.302 & 1.312 & 0.061 & 0.985 \\
1.0 & 15519 &  94.8 &  8.70 & 25.44 & 2.324 & 1.296 & 0.064 & 0.984 \\
2.0 & 14274 &  87.2 &  9.75 & 27.83 & 2.491 & 1.216 & 0.081 & 0.971 \\
3.0 & 12785 &  78.1 & 11.29 & 32.14 & 2.676 & 1.150 & 0.101 & 0.954 \\
\bottomrule
\end{tabular}
\end{table}

\textbf{Online timing parameters adaptation.} The empirical fits above provide offline MLE estimates as ground-truth references. Starting from generic initial values, our online update rule converges to $\mu \approx 4.1$ and $\sigma \approx 1.0$ for chat, and to $\mu \approx 1.8$ and $\sigma \approx 1.1$ for agentic workloads.
These estimates recover the same order of magnitude as the offline references for both $\mu$ and $\sigma$.
The moderate underestimation of $\sigma$ is expected for an EWMA-style online estimator, which discounts older observations more aggressively than batch MLE and therefore underweights the long upper tail.
In our setting, this does not harm cache decisions because the fitted distribution is used mainly to rank cached blocks by relative recency, rather than to predict exact tail probabilities.
These results show that our online estimator can recover the dominant timing scale of each workload from runtime observations alone, without manual tuning.
The pseudo code for updating can be found in Alg.\ref{alg:lognormal_update}.


 We analyze the conversational characteristics of three representative real-world datasets, namely ShareGPT, LMSys, and Chatbot-Arena, by examining the proportions of single-turn and multi-turn conversations, average user turns, turn-count distributions, and follow-up request ratios in Fig. \ref{fig:conversation_turn_statistics}. 
 The results show substantial heterogeneity across datasets: ShareGPT contains 75\% multi-turn conversations with an average of 3.6 turns, LMSys contains 33\% multi-turn conversations with an average of 2.0 turns, while Chatbot-Arena is dominated by single-turn interactions, with only 12\% multi-turn conversations and an average of 1.2 turns. In particular, ShareGPT exhibits a significant proportion of long conversations, where 29\% of sessions contain more than five turns, highlighting the high potential for KV cache reuse in long-context interactive workloads and motivating adaptive cache eviction strategies for heterogeneous real-world scenarios.

\begin{figure}[!ht]
    \centering
    \includegraphics[width=\linewidth]{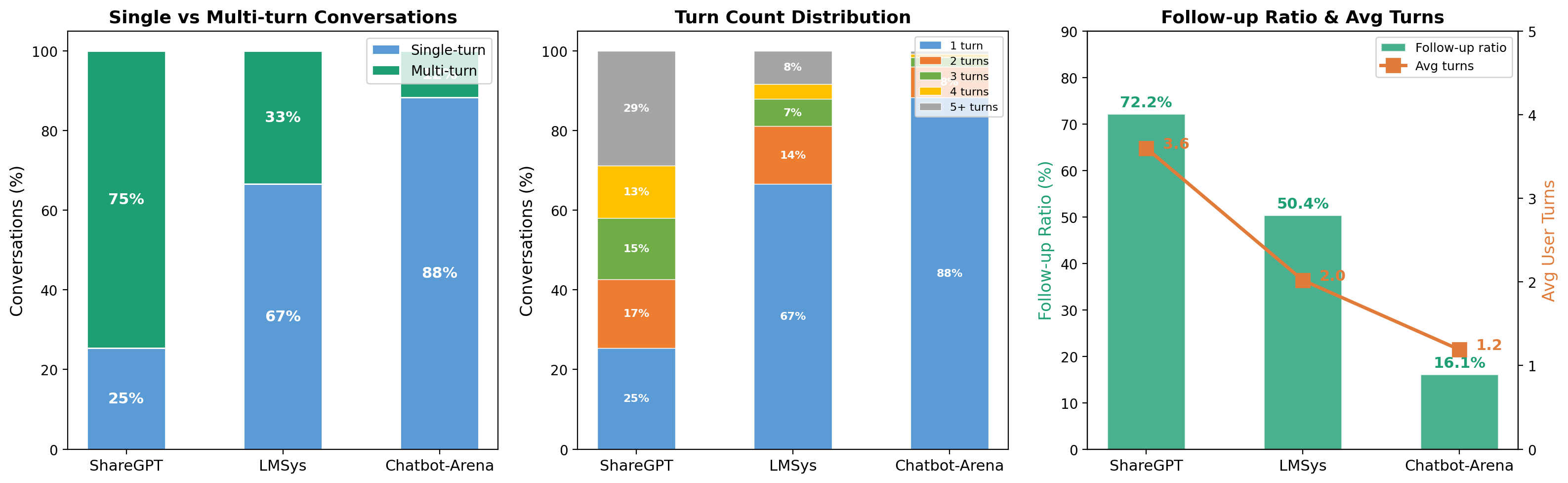}
    \caption{Conversation turn statistics across ShareGPT, LMSys, and Chatbot-Arena.}
    \label{fig:conversation_turn_statistics}
\end{figure}

We train a lightweight continuation classifier on real-world conversational data to predict whether a user will continue the interaction after a given \textless prompt, response\textgreater{} pair.
As shown in this Fig. \ref{fig:continuation_classifier_training}, the training loss consistently decreases from nearly $6.0$ to approximately $4.1$, while the training accuracy steadily improves from about $73\%$ to over $82\%$ across 50 epochs, demonstrating stable convergence and effective learning of conversational continuation patterns. These results indicate that semantic features extracted from the current interaction can effectively capture user follow-up behaviors, enabling adaptive cache management for heterogeneous multi-turn workloads.

\begin{figure}[!ht]
    \centering
    \includegraphics[width=0.95\linewidth]{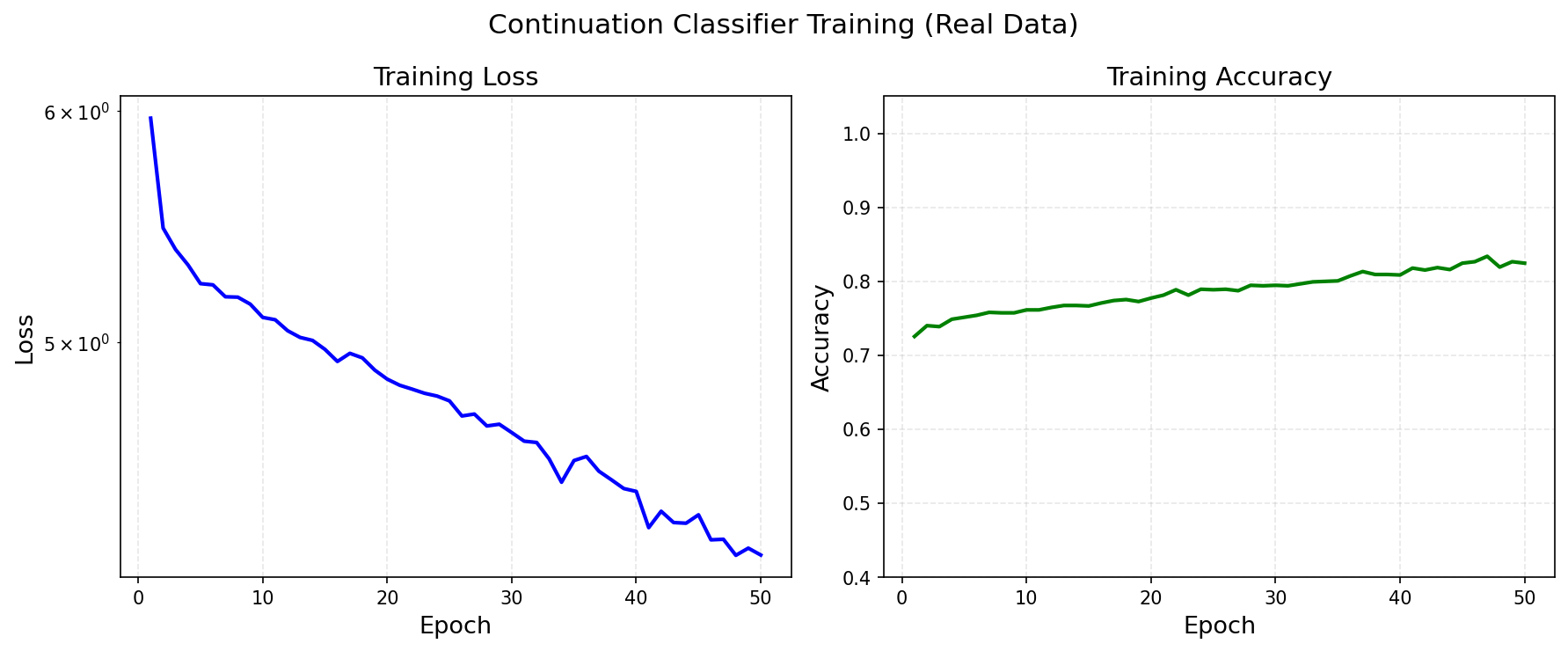}
    \caption{Continuation classifier training curves on real data. }
    \label{fig:continuation_classifier_training}
\end{figure}


\begin{figure}[H]
    \centering 
    \begin{minipage}[b]{0.45\textwidth} 
    \centering 
    \includegraphics[width=\linewidth]{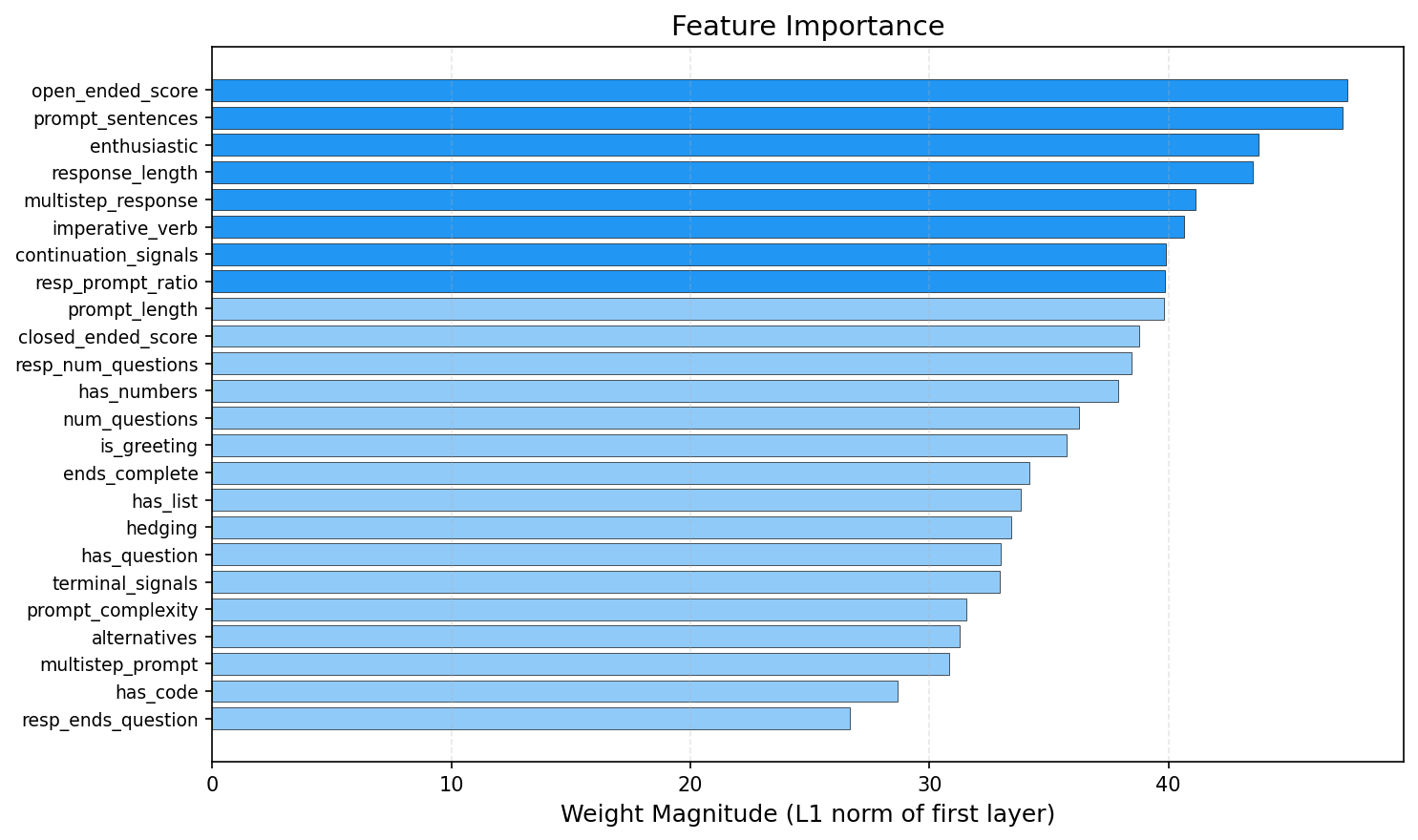}
    \caption{Feature importance of the continuation classifier.}
    \label{fig:continuation_feature_importance}
    \end{minipage}
    \begin{minipage}[b]{0.45\textwidth} 
    \centering 
    \includegraphics[width=\linewidth]{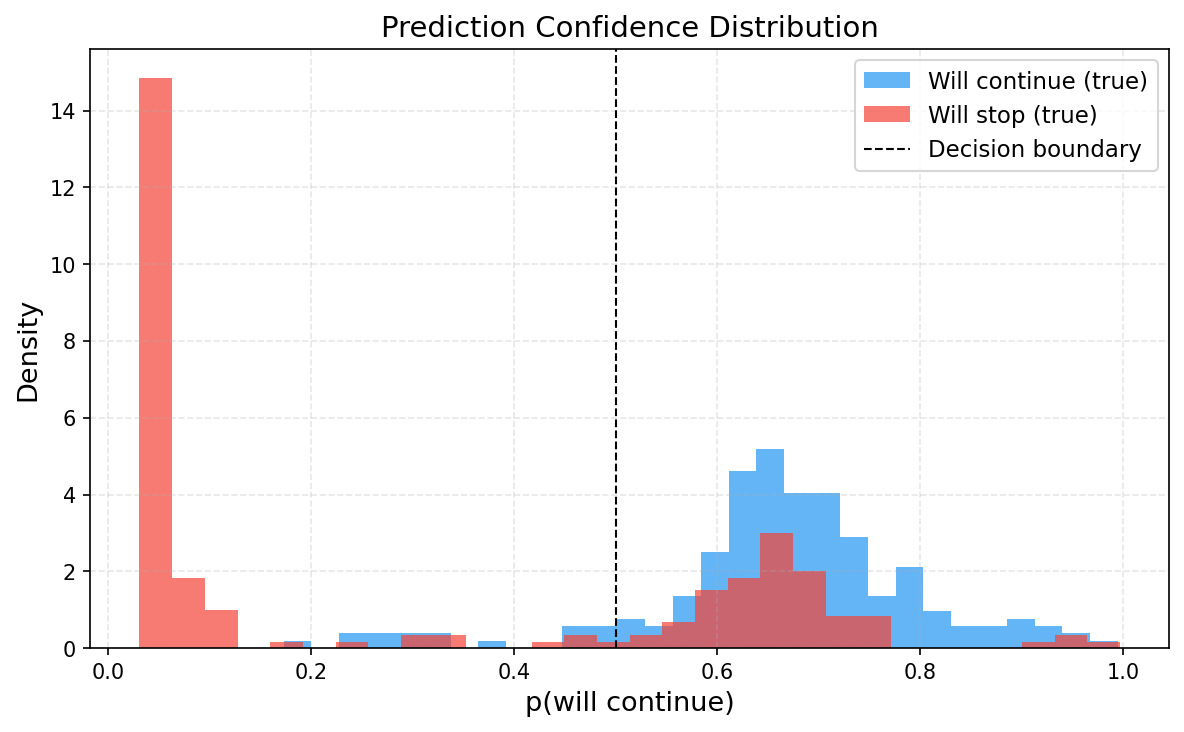}
    \caption{Prediction confidence distribution of the continuation classifier. }
    \label{fig:continuation_confidence_distribution}
    \end{minipage}
\end{figure}

We studied the importance of each feature in classifier training for the predictor in Fig. \ref{fig:continuation_feature_importance}.
It presents the feature importance analysis of the continuation classifier. The most influential features are \textit{open\_ended\_score} and \textit{prompt\_sentences}, indicating that the complexity and openness of the user prompt are the strongest signals for predicting future continuation behavior. Several response-side features, including \textit{enthusiastic}, \textit{response\_length}, and \textit{multistep\_response}, also exhibit high importance, suggesting that the interactivity and engagement level of the assistant response significantly affect whether users continue the conversation. Interestingly, features such as \textit{has\_code} and \textit{resp\_ends\_question} contribute relatively little, implying that simple surface-level response patterns are less predictive than deeper semantic interaction signals.


We also investigated the prediction confidence distribution of the continuation classifier in Fig. \ref{fig:continuation_confidence_distribution}. 
It illustrates the prediction confidence distribution of the continuation classifier. Most ``will stop'' samples are concentrated near $p \approx 0.05$, indicating that the model can confidently identify strong conversational termination signals, while ``will continue'' samples are distributed more broadly between $0.5$ and $0.9$ and partially overlap with stopping cases. This overlap suggests that continuation prediction inherently contains uncertainty, since whether a user continues the interaction often depends not only on the semantic content of the current \textless prompt, response\textgreater{} pair, but also on external and unpredictable user behaviors beyond the conversation itself.


\begin{figure}[H]
    \centering 
    \begin{minipage}[b]{0.45\textwidth} 
    \centering 
    \includegraphics[width=\linewidth]{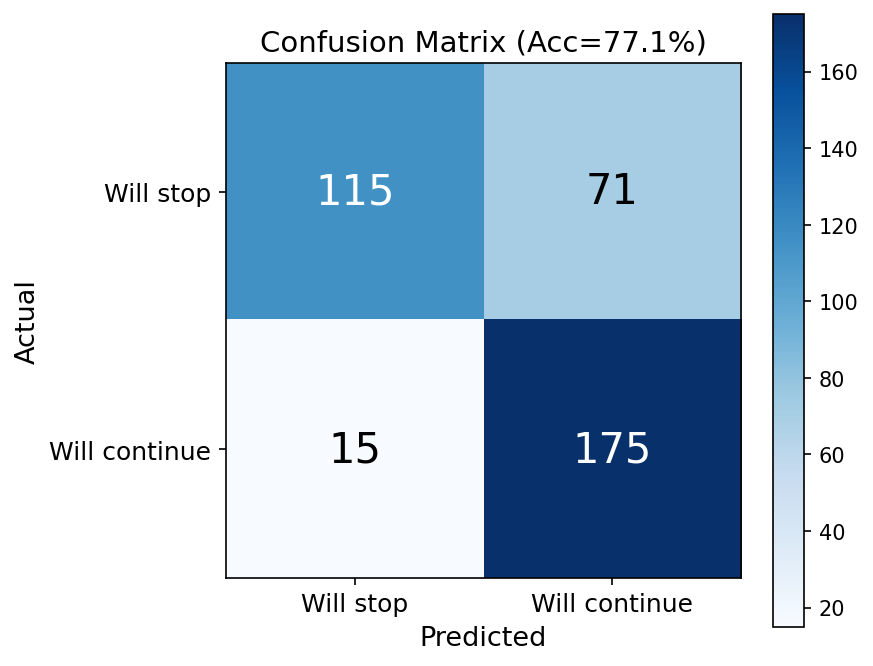}
    \caption{Continuation classifier confusion matrix. }
    \label{fig:continuation_classifier_confusion_matrix}
    \end{minipage}
    \begin{minipage}[b]{0.45\textwidth} 
    \centering 
    \includegraphics[width=\linewidth]{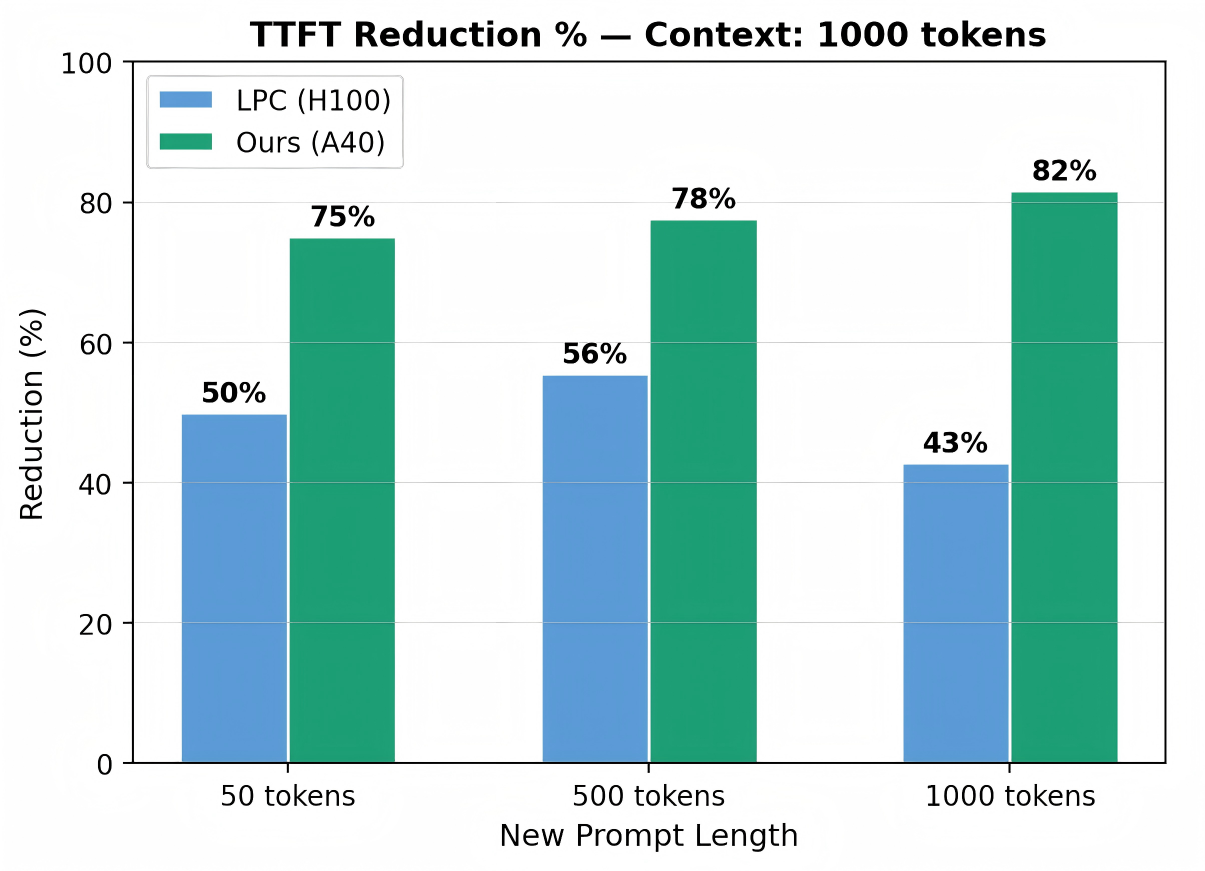}
    \caption{Comparison between LPC and our method under long context.}
    \label{fig:lpc_comparison_context_scaling}
    \end{minipage}
\end{figure}

To better interpret the prediction outcomes, we show the continuation classifier confusion matrix in Fig. \ref{fig:continuation_classifier_confusion_matrix}
Our Figure presents the confusion matrix of the continuation classifier, achieving an overall accuracy of $77.1\%$ and an F1-score of $0.803$, indicating that predicting user continuation behavior from conversational semantics is a realistic and non-trivial task. The classifier correctly identifies 175 continuation cases and 115 stopping cases, while producing substantially more false positives ($71$) than false negatives ($15$). This asymmetry is desirable for KV cache management, since falsely retaining several unnecessary cache blocks incurs only moderate memory overhead, whereas incorrectly evicting sessions that will actually continue can significantly reduce future cache reuse opportunities and increase recomputation cost.


To fully explain the scalability of our method, we integrate our method with the LLAMA-8B model and compare it with the LPC implemented on Qwen3-32B.
Specifically, we compare them on the TTFT in the long context case in Fig. \ref{fig:lpc_comparison_context_scaling}.
We show that our method exhibits a clear improvement over the TTFT reduction in the 1000-token context case, which demonstrates that our method is not only plug-and-play but also effective in the long context scenarios.
This also indicates that our eviction policy can preserve the token with high reuse potentials and evict only the low-value blocks, which further validates the effectiveness of our method.


\clearpage
\section*{NeurIPS Paper Checklist}

The checklist is designed to encourage best practices for responsible machine learning research, addressing issues of reproducibility, transparency, research ethics, and societal impact. Do not remove the checklist: {\bf The papers not including the checklist will be desk rejected.} The checklist should follow the references and follow the (optional) supplemental material.  The checklist does NOT count towards the page
limit.

Please read the checklist guidelines carefully for information on how to answer these questions. For each question in the checklist:
\begin{itemize}
    \item You should answer \answerYes{}, \answerNo{}, or \answerNA{}.
    \item \answerNA{} means either that the question is Not Applicable for that particular paper or the relevant information is Not Available.
    \item Please provide a short (1--2 sentence) justification right after your answer (even for \answerNA).
\end{itemize}

{\bf The checklist answers are an integral part of your paper submission.} They are visible to the reviewers, area chairs, senior area chairs, and ethics reviewers. You will also be asked to include it (after eventual revisions) with the final version of your paper, and its final version will be published with the paper.

The reviewers of your paper will be asked to use the checklist as one of the factors in their evaluation. While \answerYes{} is generally preferable to \answerNo{}, it is perfectly acceptable to answer \answerNo{} provided a proper justification is given (e.g., error bars are not reported because it would be too computationally expensive'' or ``we were unable to find the license for the dataset we used''). In general, answering \answerNo{} or \answerNA{} is not grounds for rejection. While the questions are phrased in a binary way, we acknowledge that the true answer is often more nuanced, so please just use your best judgment and write a justification to elaborate. All supporting evidence can appear either in the main paper or the supplemental material, provided in appendix. If you answer \answerYes{} to a question, in the justification please point to the section(s) where related material for the question can be found.

\begin{enumerate}

\item {\bf Claims}
    \item[] Question: Do the main claims made in the abstract and introduction accurately reflect the paper's contributions and scope?
    \item[] Answer: \answerYes{}
    \item[] Justification: The abstract and Section~1 enumerate four contributions (empirical characterization of token-type and session-structure reuse, the SAECache multi-queue eviction policy, online token-type-aware weighting, and end-to-end TTFT evaluation), all of which are validated in Sections~2--4 and Appendices~A--E.

\item {\bf Limitations}
    \item[] Question: Does the paper discuss the limitations of the work performed by the authors?
    \item[] Answer: \answerYes{}
    \item[] Justification: Limitations are discussed inline: Section~4 reports a $12$--$34\%$ TTFT regression on Chatbot-Arena (single-turn-dominated workload) where multi-queue overhead exceeds the prefill savings, and Appendix~B documents the EWMA-style underestimation of $\sigma$ in heavy-tail regimes.

\item {\bf Theory assumptions and proofs}
    \item[] Question: For each theoretical result, does the paper provide the full set of assumptions and a complete (and correct) proof?
    \item[] Answer: \answerNA{}
    \item[] Justification: The paper does not include formal theoretical results (no theorems, lemmas, or formal proofs); all derivations are empirical characterizations and online learning rules.

    \item {\bf Experimental result reproducibility}
    \item[] Question: Does the paper fully disclose all the information needed to reproduce the main experimental results of the paper to the extent that it affects the main claims and/or conclusions of the paper (regardless of whether the code and data are provided or not)?
    \item[] Answer: \answerYes{}
    \item[] Justification: Section~3 and Algorithms~1--4 specify the eviction procedure and online update rules; Section~4 describes the baselines, datasets, hardware (NVIDIA A40, vLLM v0.8.5, Qwen2.5-1.5B-Instruct), and request injection intervals; Appendix~B documents all parameter learning rules and clamp ranges; Appendix~C reports a 30-configuration hyperparameter sensitivity sweep.

\item {\bf Open access to data and code}
    \item[] Question: Does the paper provide open access to the data and code, with sufficient instructions to faithfully reproduce the main experimental results, as described in supplemental material?
    \item[] Answer: \answerYes{}
    \item[] Justification: Code is released at submission to preserve anonymity; the evaluation traces (ShareGPT, LMSys, Chatbot-Arena, Qwen-Bailian, CC-Bench-V1.1) are released datasets accessible from their original providers, and Algorithms~1--4 plus Appendix~B provide full pseudocode and parameter update rules sufficient for re-implementation. The code will be released upon acceptance.
    \item[] Guidelines:
    \begin{itemize}
        \item The answer \answerNA{} means that paper does not include experiments requiring code.
        \item Please see the NeurIPS code and data submission guidelines (\url{https://nips.cc/public/guides/CodeSubmissionPolicy}) for more details.
        \item While we encourage the release of code and data, we understand that this might not be possible, so \answerNo{} is an acceptable answer. Papers cannot be rejected simply for not including code, unless this is central to the contribution (e.g., for a new open-source benchmark).
        \item The instructions should contain the exact command and environment needed to run to reproduce the results. See the NeurIPS code and data submission guidelines (\url{https://nips.cc/public/guides/CodeSubmissionPolicy}) for more details.
        \item The authors should provide instructions on data access and preparation, including how to access the raw data, preprocessed data, intermediate data, and generated data, etc.
        \item The authors should provide scripts to reproduce all experimental results for the new proposed method and baselines. If only a subset of experiments are reproducible, they should state which ones are omitted from the script and why.
        \item At submission time, to preserve anonymity, the authors should release anonymized versions (if applicable).
        \item Providing as much information as possible in supplemental material (appended to the paper) is recommended, but including URLs to data and code is permitted.
    \end{itemize}

\item {\bf Experimental setting/details}
    \item[] Question: Does the paper specify all the training and test details (e.g., data splits, hyperparameters, how they were chosen, type of optimizer) necessary to understand the results?
    \item[] Answer: \answerYes{}
    \item[] Justification: Section~4 reports baselines (LRU, LPC, Token-Weight-Only, Fixed-Param Multi-Queue), datasets and their multi-turn ratios ($74\%$/$33\%$/$12\%$), and the four request injection intervals ($\{0.02, 0.03, 0.05, 0.08\}$~s); Appendix~B specifies all online update rules with clamp ranges, EMA factors, and sample thresholds; Appendix~C sweeps $\alpha \in \{0.5,1,2,5,10,20\}$ and $\beta \in \{0.5,1,2,5,10\}$.

\item {\bf Experiment statistical significance}
    \item[] Question: Does the paper report error bars suitably and correctly defined or other appropriate information about the statistical significance of the experiments?
    \item[] Answer: \answerNo{}
    \item[] Justification: The main TTFT and hit-ratio results in Figures~\ref{fig:ttft} and~\ref{fig:hitrate} are reported as point estimates without error bars; the hyperparameter sensitivity sweep in Appendix~C provides robustness across $30$ $(\alpha, \beta)$ configurations on the balanced workload, but per-run variability across multiple random seeds is not reported.

\item {\bf Experiments compute resources}
    \item[] Question: For each experiment, does the paper provide sufficient information on the computer resources (type of compute workers, memory, time of execution) needed to reproduce the experiments?
    \item[] Answer: \answerYes{}
    \item[] Justification: Section~4 specifies the hardware (a single NVIDIA A40 GPU) and software stack (vLLM v0.8.5 with the Qwen2.5-1.5B-Instruct serving model); experiments are trace-driven simulations under controlled request injection intervals, and the eviction policy itself adds only constant per-eviction overhead aside from candidate selection. Per-run wall-clock timings and total project compute are not separately reported, as the dominant cost is the trace-replay rather than model training.

\item {\bf Code of ethics}
    \item[] Question: Does the research conducted in the paper conform, in every respect, with the NeurIPS Code of Ethics \url{https://neurips.cc/public/EthicsGuidelines}?
    \item[] Answer: \answerYes{}
    \item[] Justification: We have reviewed the NeurIPS Code of Ethics and the research described in this paper conforms with it in every respect.

\item {\bf Broader impacts}
    \item[] Question: Does the paper discuss both potential positive societal impacts and negative societal impacts of the work performed?
    \item[] Answer: \answerYes{}
    \item[] Justification: SAECache reduces redundant prefill computation in LLM serving and yields $1.4\times$--$2.7\times$ TTFT improvement, which translates into lower inference energy consumption and improved latency for users on resource-constrained deployments; we are not aware of direct negative societal applications because the contribution is a foundational systems-level cache eviction policy with no model-release component.

\item {\bf Safeguards}
    \item[] Question: Does the paper describe safeguards that have been put in place for responsible release of data or models that have a high risk for misuse (e.g., pre-trained language models, image generators, or scraped datasets)?
    \item[] Answer: \answerNA{}
    \item[] Justification: The paper proposes a cache eviction policy and does not release pre-trained generative models, image generators, or scraped datasets that pose misuse risks.

\item {\bf Licenses for existing assets}
    \item[] Question: Are the creators or original owners of assets (e.g., code, data, models), used in the paper, properly credited and are the license and terms of use explicitly mentioned and properly respected?
    \item[] Answer: \answerYes{}
    \item[] Justification: All evaluation datasets (ShareGPT, LMSys, Chatbot-Arena cited in Section~4; Qwen-Bailian and CC-Bench-V1.1 cited in Appendix~E), the backbone serving model (Qwen2.5-1.5B-Instruct, Section~4), the vLLM framework, and the LPC predecessor (Appendix~A) are properly attributed to their original creators as standard public research assets used under their original terms.

\item {\bf New assets}
    \item[] Question: Are new assets introduced in the paper well documented and is the documentation provided alongside the assets?
    \item[] Answer: \answerNA{}
    \item[] Justification: The paper does not release new datasets, models, or reusable assets; the auxiliary multi-turn session predictor MLP is described as a small in-system component rather than a released artifact.

\item {\bf Crowdsourcing and research with human subjects}
    \item[] Question: For crowdsourcing experiments and research with human subjects, does the paper include the full text of instructions given to participants and screenshots, if applicable, as well as details about compensation (if any)?
    \item[] Answer: \answerNA{}
    \item[] Justification: The paper does not involve crowdsourcing or research with human subjects; all evaluation is performed on existing public conversation traces.

\item {\bf Institutional review board (IRB) approvals or equivalent for research with human subjects}
    \item[] Question: Does the paper describe potential risks incurred by study participants, whether such risks were disclosed to the subjects, and whether Institutional Review Board (IRB) approvals (or an equivalent approval/review based on the requirements of your country or institution) were obtained?
    \item[] Answer: \answerNA{}
    \item[] Justification: The paper does not involve research with human subjects; IRB or equivalent review is therefore not applicable.

\item {\bf Declaration of LLM usage}
    \item[] Question: Does the paper describe the usage of LLMs if it is an important, original, or non-standard component of the core methods in this research? Note that if the LLM is used only for writing, editing, or formatting purposes and does \emph{not} impact the core methodology, scientific rigor, or originality of the research, declaration is not required.
    \item[] Answer: \answerYes{}
    \item[] Justification: LLM is used only for language polishing during writing; the core method (multi-queue eviction policy with online weight learning) does not use LLMs as a research component, and the serving LLM (Qwen2.5-1.5B-Instruct) is the target system whose KV cache is being managed rather than an LLM used to develop the method.

\end{enumerate}

\end{document}